\documentclass[lettersize,journal]{IEEEtran}
\usepackage{amsmath,amsfonts}
\usepackage{algorithmic}
\usepackage{algorithm}
\usepackage{array}
\usepackage[caption=false,font=normalsize,labelfont=sf,textfont=sf]{subfig}
\usepackage{textcomp}
\usepackage{stfloats}
\usepackage{url}
\usepackage{verbatim}
\usepackage{graphicx}
\usepackage{cite}
\usepackage{lineno,hyperref}

\usepackage{xcolor}
\usepackage{multirow}
\usepackage{amsmath,amsthm,amsfonts,amssymb,bm}
\usepackage{booktabs}
\usepackage{makecell}

\usepackage{xcolor}
\definecolor{b}{RGB}{0,0,250}

\begin{document}

\title{How Vision-Language Tasks Benefit from Large Pre-trained Models: A Survey}

\author{Yayun~Qi,~Hongxi~Li,~Yiqi~Song,~Xinxiao~Wu,~\IEEEmembership{Member,~IEEE,}~Jiebo~Luo,~\IEEEmembership{Fellow,~IEEE}%
\thanks{Yayun Qi, Hongxi Li, and Yiqi Song are with the Beijing Laboratory of Intelligent Information Technology, School of Computer Science, Beijing Institute of Technology, Beijing 100081, China. Emails: \{qiyayun,lihongxi,yiqis\}@bit.edu.cn}
\thanks{Xinxiao Wu is with the Beijing Laboratory of Intelligent Information Technology, School of Computer Science, Beijing Institute of Technology, Beijing 100081, China, and also with the Guangdong Provincial Laboratory of Machine Perception and Intelligent Computing, Shenzhen MSU-BIT University, Shenzhen 518172, China. Email: wuxinxiao@bit.edu.cn}
\thanks{Jiebo Luo is with the Department of Computer Science, University of Rochester, Rochester, NY 14627 USA. Email: jiebo.luo@gmail.com}
\thanks{Xinxiao Wu is the corresponding author.}}

\markboth{Journal of IEEE,~Vol.~14, No.~8, August~2021}%
{Shell \MakeLowercase{\textit{et al.}}: A Sample Article Using IEEEtran.cls for IEEE Journals}

\IEEEpubid{0000--0000/00\$00.00~\copyright~2024 IEEE}

\maketitle

\begin{abstract}
The exploration of various vision-language tasks, such as visual captioning, visual question answering, and visual commonsense reasoning, is an important area in artificial intelligence and continuously attracts the research community's attention. Despite the improvements in overall performance, classic challenges still exist in vision-language tasks and hinder the development of this area. In recent years, the rise of pre-trained models is driving the research on vision-language tasks. Thanks to the massive scale of training data and model parameters, pre-trained models have exhibited excellent performance in numerous downstream tasks. Inspired by the powerful capabilities of pre-trained models, new paradigms have emerged to solve the classic challenges. Such methods have become mainstream in current research with increasing attention and rapid advances. In this paper, we present a comprehensive overview of how vision-language tasks benefit from pre-trained models. First, we review several main challenges in vision-language tasks and discuss the limitations of previous solutions before the era of pre-training. Next, we summarize the recent advances in incorporating pre-trained models to address the challenges in vision-language tasks. Finally, we analyze the potential risks associated with the inherent limitations of pre-trained models and discuss possible solutions, attempting to provide future research directions.  

\end{abstract}

\begin{IEEEkeywords}
Vision-language tasks, Pre-trained model, Vision-language model, Large language model.
\end{IEEEkeywords}

\section{Introduction}

\begin{figure}[!t]
\centering
\includegraphics[width=0.85\linewidth]{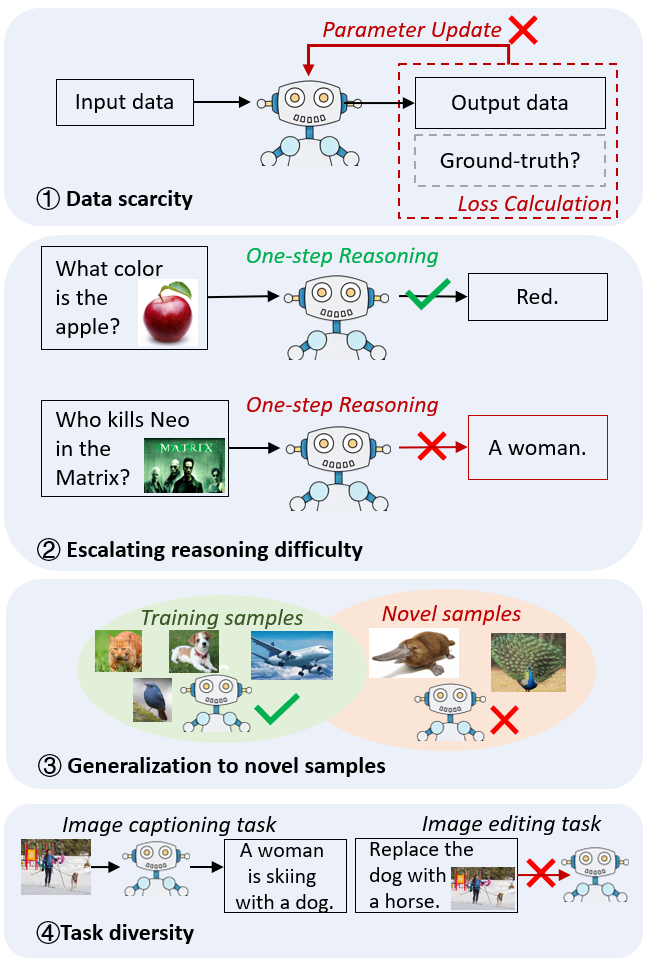}
\caption{An illustration of four classic challenges in vision-language tasks.}
\label{fig0}
\end{figure}
\IEEEPARstart{A}{s} an intersection of computer vision and natural language processing, vision-language tasks have attracted significant attention in recent years~\cite{uppal2022multimodal}.  Research in this area aims to bridge the gap between visual and textual modalities, offering promising enhancements for downstream applications~\cite{zhou2023vision}. 
The scope of visual-language tasks is quite broad, encompassing classic tasks such as visual captioning~\cite{anderson2018bottom,lin2022swinbert}, visual question answering~\cite{antol2015vqa,tapaswi2016movieqa}, visual-text retrieval~\cite{zhang2020context,chen2020fine}, and so on. 
In addition, some traditional visual recognition tasks ~\cite{menon2022visual,yan2023learning,dai2023exploring,ren2024chatgpt} (e.g. image classification) begin to emphasize the semantic meaning behind the language supervision, rather than simply treating these signals as one-hot vectors. By incorporating semantic information as an external language modality, these tasks can also be viewed as visual-language tasks~\cite{gan2022vision}. As research in the field of vision-language continues to advance, tasks are now addressing a broader scope of visual data, expanding from images~\cite{vinyals2015show,antol2015vqa,yan2015deep,gregor2015draw} to videos~\cite{pei2019memory,fan2019heterogeneous,tulyakov2018mocogan,chen2020rethinking}. 
Furthermore, the capabilities required by these tasks have shifted from basic perception~\cite{zou2023object,zellers2018neural} to more advanced reasoning~\cite{zellers2019recognition,xie2019visual,hessel2022abduction,bitton2023breaking,zhong2024let}. These advancements have resulted in significant improvements in performance across various aspects. 

\IEEEpubidadjcol

Despite the remarkable advances in many vision-language tasks, some classic challenges continue to hinder further improvement in this area and negatively impact practical applications. In Fig.~\ref{fig0}, we present four main challenges faced by vision-language tasks. 
From the perspective of training data, manually annotating data for vision-language tasks is labor-intensive, and it is challenging to automate this by masking  data to construct self-supervised training data like uni-modal tasks, which results in a significant scarcity of annotated training data.  
From the perspective of reasoning, the difficulties of both understanding the visual content and answering the questions have increased dramatically, requiring models to perform multi-hop reasoning rather than directly obtaining answers from visual information, which poses challenges to complex reasoning. 
From the perspective of generalization, when encountering new samples that exceed the model's capabilities in real-world scenarios, the limited cross-modal knowledge in the training set is not sufficient to  effectively handle these samples. 
From the perspective of task diversity, different tasks involve different reasoning processes and input-output workflows, and require different models, one of which is  tailored to a specific training process and a unique set of parameters, resulting in poor model universality. Researchers continuously propose methods to address or alleviate these challenges, and new solutions to these challenges have emerged thanks to the rapid development of large-scale pre-trained models.

With the rise of pre-trained Large Language Models (LLMs), the approach to performing downstream tasks has gradually shifted from neural network-based fully supervised learning to the ``pre-train \& fine-tune'' paradigm and the ``pre-train, prompt, predict'' paradigm~\cite{liu2023pre}. In this process, numerous LLMs have emerged (\textit{e.g.} LLaMA~\cite{touvron2023llama}, Vicuna~\cite{vicuna2023}, QWEN~\cite{bai2023qwen}), and the scale of training data and model parameters has continued to increase. Consequently, these models have exhibited intelligent emergence and performed well in many downstream tasks of NLP.  
Inspired by the success of LLMs, the vision-language field has also seen the emergence of discriminative pre-trained Vison-Language Models (VLMs) represented by CLIP~\cite{radford2021learning} (\textit{e.g.} ALIGN~\cite{jia2021scaling}, VLMo~\cite{bao2022vlmo}, FLIP~\cite{yao2021filip}), and generative VLMs represented by BLIP~\cite{li2022blip} (\textit{e.g.} MiniGPT-4~\cite{zhu2023minigpt}, LLaVA~\cite{liu2023llava}, Video-LLaMA~\cite{zhang2023video}). Compared to LLMs, VLMs  incorporate additional comprehensive knowledge of vision-language correspondence.

Thanks to the outstanding few-shot learning capabilities, extensive knowledge storage, and robust generalization of large pre-trained models, more and more methods of vision-language tasks are integrating pre-trained models to benefit from these models. 
The motivations and strategies for using pre-trained models vary significantly among these methods. 
However, the research community lacks a comprehensive survey that systematically summarizes these works. 
Such a survey is crucial for reviewing and categorizing existing methods, providing valuable insights for future research and offering researchers a coherent overview.

In this paper, we present a comprehensive overview of how vision-language tasks benefit from large pre-trained models. To the best of our  knowledge, this survey is pioneering in its focus on this topic and in categorizing methods according to the challenges they tackle.
We believe this survey can provide valuable guidance for researchers to deepen their understanding of this field and promote further developments in the community. 

To ensure the systematic nature of this paper, we first review several main challenges encountered in vision-language tasks. In addition, we discuss the limitations of  solutions before the era of pre-trained models. Subsequently, we  summarize existing methods that use the capabilities of pre-trained models to address the challenges in vision-language tasks. These methods are categorized according to the specific challenges they target and further subdivided into detailed classifications according to their paradigms. Notably, we encapsulate these methodological paradigms into illustrations for a more intuitive understanding. 
The tasks involved include visual classification, visual captioning, visual question answering, image editing, video localization, visual data generation, open-vocabulary object detection, and so on. Finally, we also explore the potential risks associated with the use of pre-trained models in vision-language tasks due to the inherent limitations of pre-trained models, and discuss potential solutions to guide future research directions. 

The remainder of the paper is organized as follows. Section~\ref{sec:prob} introduces the main challenges encountered in vision-language tasks. Section~\ref{sec:method} reviews the solutions to the above-mentioned challenges using pre-trained models. Section~\ref{sec:risk} discusses the potential risks associated with integrating pre-trained models and explores promising future directions to mitigate these risks.

\section{Related Surveys and Differences}
With the advancement of research on pre-trained models, there have been several surveys~\cite{zhao2023survey,naveed2023comprehensive,chang2024survey,wang2023large,awais2023foundational,yin2023survey,zhang2024mm,wang2024exploring,tang2023video} that focus on summarizing the fundamental information of pre-trained models (\textit{i.e.} how they work), such as their training data, model structures, evaluation practices, and pre-training objectives and strategies. Among them, the VLM-related surveys~\cite{wang2023large,awais2023foundational,yin2023survey,zhang2024mm,wang2024exploring,tang2023video} aim to cover current VLMs and classify them using different taxonomies, in order to comprehensively review the unique feature of each VLM and inspire future research on designing novel VLMs.

Another trend in surveys~\cite{zhang2024vision,he2024llms,miyai2024generalized} on pre-trained models is to summarize methods that combine pre-trained models to tackle specific tasks (\textit{i.e.} how to use). 
These surveys typically categorize methods based on their paradigms or target tasks to inspire future research to apply or improve existing paradigms on various tasks.  
However, surveys on how to use pre-trained models are scarce and mostly focus on limited pre-trained models and modalities. The research community is short of a survey that covers the integration of both LLMs and VLMs to tackle vision-language tasks involving images, videos, and text. To this end, we present a survey that focuses on the novel paradigms of incorporating pre-trained models to address classic challenges in vision-language tasks. Table I shows the differences between our survey and the existing related surveys in terms of content and coverage. Table~\ref{tab:survey} shows the distinctions between our survey and other surveys in terms of content and coverage. 

\begin{table*}[htbp]
  \centering
  \caption{Comparison between our survey and  other surveys. We divide these surveys into two classes according to their main content.}
  \resizebox{0.95\textwidth}{!}{
    \begin{tabular}{l|p{21em}|c|c|c|c|c|p{14.165em}}
    \toprule
    \multicolumn{1}{c|}{\multirow{2}[3]{*}{Main Content}} & \multicolumn{1}{c|}{\multirow{2}[3]{*}{Survey}} & \multicolumn{2}{c|}{Pre-trained Model} & \multicolumn{3}{c|}{Modality} & \multicolumn{1}{c}{\multirow{2}[3]{*}{Taxonomy}} \\
\cmidrule{3-7}    \multicolumn{1}{c|}{} &       & \ LLM \  & VLM & Image & Video & Text &  \\
    \midrule
    \multicolumn{1}{l|}{\multirow{5}[30]{*}{\makecell[l]{Summarize funda-\\mental information \\of pre-trained \\models}}} & Foundational Models Defining a New Era in Vision: A Survey and Outlook~\cite{awais2023foundational} &       & $\surd$ & $\surd$ & $\surd$ & $\surd$ & Categorize VLMs based on the architectures \\
    \cmidrule{2-8}
          & Video Understanding with Large Language Models: A Survey~\cite{tang2023video} & $\surd$ & $\surd$ &       & $\surd$ & $\surd$ & Categorize VLMs based on the strategies to integrate LLMs \\
    \cmidrule{2-8}
          & Exploring the Reasoning Abilities of Multimodal Large Language Models (MLLMs): A Comprehensive Survey on Emerging Trends in Multimodal Reasoning~\cite{wang2024exploring}  &       & $\surd$ & $\surd$ & $\surd$ & $\surd$ & Categorize VLMs based on the strategies to improve reasoning abilities \\
    \cmidrule{2-8}
          & Large-scale Multi-Modal Pre-trained Models: A Comprehensive Survey~\cite{wang2023large} &       & $\surd$ & $\surd$ & $\surd$ & $\surd$ & Categorize VLMs based on the pre-training strategies \\
    \cmidrule{2-8}
          & MM-LLMs: Recent Advances in MultiModal Large Language Models~\cite{zhang2024mm} &       & $\surd$ & $\surd$ & $\surd$ & $\surd$ & Categorize VLMs from both functional and design perspectives \\
    \midrule
    \multicolumn{1}{l|}{\multirow{4}[13]{*}{\makecell[l]{Summarize the \\integration of \\pre-trained models \\in specific tasks}}} & Vision-Language Models for Vision Tasks: A Survey~\cite{zhang2024vision} &       & $\surd$ & $\surd$ & $\surd$ &       & Categorize methods based on the paradigms of incorporating VLMs \\
         \cmidrule{2-8}
          & LLMs Meet Multimodal Generation and Editing: A Survey~\cite{he2024llms} & $\surd$ &       & $\surd$ & $\surd$ & $\surd$ & Categorize methods based on the target tasks \\
          \cmidrule{2-8}
          & Generalized Out-of-Distribution Detection and Beyond in Vision Language Model Era: A Survey~\cite{miyai2024generalized} &       & $\surd$ & $\surd$ & & & Categorize methods based on the target tasks   \\  
          \cmidrule{2-8}
          & {How Vision-Language Tasks Benefit from Large Pre-trained Models: A Survey (Ours)}  & $\surd$ & $\surd$ & $\surd$ & $\surd$ & $\surd$ & Categorize methods based on the target challenges \\
    \bottomrule
    \end{tabular}}%
  \label{tab:survey}%
\end{table*}%

\section{Challenges in Vision-language Tasks}
\label{sec:prob}
In this section, we introduce four major challenges faced by all models in vision-language tasks, including data scarcity, escalating reasoning complexity, generalization to novel samples, and task diversity. For each challenge, we discuss the corresponding methods and their limitations before the era of pre-training. 

\subsection{Data Scarcity}

For supervised learning methods of vision-language tasks, such as image captioning and visual question answering, annotated data plays a crucial role in their training process since the annotated data is the foundation for parameter update. However, manually annotating these huge data is quite laborious, and collecting annotated data for certain tasks such as counterfactual image editing and counter-intuitive reasoning from real-world scenarios is particularly challenging. On the other hand, the training data for vision-language tasks involve two different modalities, which makes it impractical to construct self-supervised data using unlabeled uni-modal data via proxy tasks, as commonly done in natural language understanding and generation~\cite{devlin2018bert,sun2019ernie,brown2020language}. 
These circumstances lead to the challenge of data scarcity in training, and the primary concern is how to accomplish vision-language tasks without  available annotated data.

To address this challenge in vision-language tasks, researchers have developed semi-supervised learning methods~\cite{kim2019image,mandal2019semi} that train models using a small set of labeled visual-textual data while simultaneously using a large set of unlabeled text and visual data to enhance generalization. Meanwhile, weakly supervised learning methods~\cite{duan2018weakly,luo2021weakly} have been proposed to learn from incomplete visual-textual annotations, noisy labels, or weak supervision signals. For example, a VQA model can be trained with only image-caption pairs as supervision. In addition, unsupervised learning methods~\cite{feng2019unsupervised,laina2019towards} have been designed to discover patterns and relationships between visual and textual data from unpaired visual and text inputs, using techniques like contrastive and adversarial learning. Despite these advances, as the amount of annotated visual-textual data decreases, these methods often struggle to achieve performance comparable to fully supervised learning. Moreover, in semi-supervised and weakly supervised settings, limited annotations often cause models to overfit the observed visual-textual patterns, and this overfitting is subsequently propagated when assigning pseudo labels. Furthermore, the quality of annotations plays a crucial role, as inaccurate or incomplete annotations may lead to incorrect learning of visual-textual correlations. 

\begin{table*}[htbp]
  \centering
  \caption{Taxonomy of methods that incorporate pre-trained models to tackle classic challenges in vision-language tasks. Each classification of methods follows a summary of methods covered in this survey and a brief description of their paradigms.}
  \resizebox{0.99\textwidth}{!}{
    \begin{tabular}{l|p{10em}|l}
    \toprule
    \multicolumn{1}{c|}{Challenge} & \multicolumn{1}{c|}{Paradigm} & \multicolumn{1}{c}{Methods} \\
    \midrule
    \multicolumn{1}{l|}{\multirow{4}[15]{*}{\makecell[l]{Data \\Scarcity}}} & \multirow{2}[8]{*}{\makecell[l]{Direct inference on\\test samples}} & \multicolumn{1}{p{40.46em}}{An LLM provides language priors and CLIP calculates image-text similarity to introduce visual constraints.~\cite{tewel2022zerocap,su2022language,zeng2023conzic,zeng2024meacap,wang2023text,tewel2022zero,jo2023zero}} \\
    \cmidrule{3-3}          & \multicolumn{1}{c|}{} & \multicolumn{1}{p{40.46em}}{A VLM converts visual information into texts and then an LLM processes these texts to complete the task.~\cite{song2022clip,yu2022towards,yang2022empirical,tiong2022plug,wang2022language,hanu2023language,bhattacharya2023video,pan2023retrieving}} \\
\cmidrule{2-3}          & Learning from unlabeled uni-modal data & \multicolumn{1}{p{40.46em}}{Use unlabeled data from the
target modality for training and then replace input features with source modality features from the common space of CLIP during testing.~\cite{nukrai2022text,gu2022can,li2023decap,wang2023association,wang2022zero,fei2023transferable,kim2023language,wang2022clip}} \\
\cmidrule{2-3}          & Generating pseudo paired data & \multicolumn{1}{p{40.46em}}{Employ generative pre-trained models to automatically
generate pseudo annotations for training or evaluation purpose.~\cite{ma2024image,liu2024improving,yang2023freemask,zhou2022towards,bitton2023breaking}} \\
    \midrule
    \multicolumn{1}{l|}{\multirow{2}[8]{*}{\makecell[l]{Escalating \\Reasoning \\Complexity}}} & Divide-and-conquer & \multicolumn{1}{p{40.46em}}{LLMs decompose the main question in a complex visual-language reasoning task into sub-questions.~\cite{chen2023see,zhou2023vicor,wang2023domino,you2023idealgpt,yang2023good,qi2023the,wang-etal-2023-filling,rajabzadeh2023multimodal}} \\
\cmidrule{2-3}          & Chain-of-Thought & \multicolumn{1}{p{40.46em}}{Decompose the direct prediction process of pre-trained models into a series of intermediate reasoning steps.~\cite{lu2022learn,zhang2023multimodal,wang2024t,mondal2024kam,chen2023measuring,zhu2023efficient,mitra2023compositional,zhang2024cocot,meng2023chain,rose2023visual,himakunthala2023let}} \\
    \midrule
    \multicolumn{1}{l|}{\multirow{2}[8]{*}{\makecell[l]{Generali-\\zation to \\Novel \\Samples}}} & Extracting semantic context from an LLM & \multicolumn{1}{p{40.46em}}{Exploit semantic context extracted from an LLM as additional cues for processing novel samples.~\cite{menon2022visual,yan2023learning,dai2023exploring,ren2024chatgpt,li2024zero,jia2023generating,yousaf2023videoprompter}} \\
\cmidrule{2-3}          & Distilling teacher knowledge from a VLM & \multicolumn{1}{p{40.46em}}{Regard a VLM as the teacher model of a close-set trained student model.~\cite{gu2022open,wu2023aligning,wang2023object,gao2022open,kuo2023open,liao2022cohoz,liao2023m,qu2024lmc}} \\
    \midrule
   \multicolumn{1}{l|}{\multirow{2}[15]{*}{\makecell[l]{Task \\Diversity}}} & Continual learning & \multicolumn{1}{p{40.46em}}{Use prompt learning or instruction tuning to enable a single VLM to continuously learn new vision-language tasks, while maintaining the performance on learned tasks.~\cite{qian2023decouple,chen2024coin,zheng2024beyond}} \\
   \cmidrule{2-3}  & Planning with natural language & \multicolumn{1}{p{40.46em}}{Treat an LLM as a planner that, given a task-related instruction, infers to call other pre-trained models for execution. The plan comes  in the form of natural language.~\cite{yang2023mm,wu2023visual,Lu2023ChameleonPC,lin2023mm}} \\
\cmidrule{2-3}          & Planning with code statements & \multicolumn{1}{p{40.46em}}{Treat an LLM as a planner that, given a task-related instruction, infers  to call other pre-trained models for execution. The plan comes in the form of code statement.~\cite{Gupta2022VisualPC,Suris2023ViperGPTVI,Choudhury2023ZeroShotVQ,Stanic2024TowardsTZ}} \\
    \bottomrule
    \end{tabular}}%
  \label{tab:method}%
\end{table*}%

\subsection{Escalating Reasoning Complexity}

In the early stages of development, vision-language tasks primarily require models to perceive and summarize visual information from given visual data. 
For instance, tasks may involve querying the location or attributes of certain entities and identifying relationships between different entities. Solutions to these tasks can often be derived directly from the visual content without  complex reasoning. 

As research progresses, more and more vision-language tasks require models to perform additional reasoning to complete. The complexity of reasoning continues to escalate since both the visual content and the associated questions become more complex than before. 
Visual scenes have evolved from simple combinations of geometric shapes~\cite{johnson2017clevr} to real scenes~\cite{lin2014microsoft} with various objects and relationships, which requires relationship reasoning. In addition, visual data now includes not only static images~\cite{plummer2015flickr30k} but also activity videos~\cite{wang2019vatex,xu2016msr} and even longer movie clips~\cite{tapaswi2016movieqa,rohrbach2015dataset} that require temporal reasoning. Furthermore, counter-intuitive images~\cite{bitton2023breaking} and humorous videos~\cite{xie2023funqa} pose challenges to models in terms of compositional and commonsense reasoning. 
The related questions shift from recognition to cognition, which  involve  exploring commonsense information and causal relationships in  visual content. This requires additional reasoning processes such as commonsense reasoning~\cite{zellers2019recognition}, abductive reasoning~\cite{hessel2022abduction,liang2022visual}, and counter-intuitive reasoning~\cite{bitton2023breaking}. In addition, visual entailment~\cite{xie2019visual} explores the logical relationships between images and sentences, requiring logical reasoning to obtain accurate answers.

Previous methods~\cite{antol2015vqa,xu2016ask} typically design different model architectures to process visual and optional text inputs, and optimize them to compute the probability distribution of answer vocabulary based on visual information. Since such single-step visual reasoning is oversimplified, these methods have difficulty extracting relevant visual features and external knowledge simultaneously, resulting in low accuracy of answers. Subsequently, some multi-step visual reasoning methods~\cite{li2022dynamic,song2018explore,wang2023knowledge} adopt a structure known as ``Retriever, Reasoner (or Reader) and Answer Predictor”. Specifically, these methods extract question-relevant visual features and knowledge, iteratively perform spatial and optional temporal reasoning on the original or decomposed questions, and predict answers. However, these methods usually limit the visual reasoning steps to two hops or require the number of hops to be set as a hyper-parameter, which lacks the flexibility and robustness to solve more complex visual reasoning problems.

\subsection{Generalization to Novel Samples}

In vision-language tasks, models typically rely on the cross-modal knowledge learned from training data to make inferences on test visual samples. However, the knowledge contained in the training data is task-specific and limited in content, and cannot cover all the information that the model may encounter in real-world scenarios. Therefore, when the trained models encounter novel visual samples that are not covered in the training data, they lack the knowledge required to make accurate visual recognition or reasoning, resulting in incorrect predictions for these novel samples. 

Under this circumstance, external knowledge plays a crucial role in supporting generalization to novel samples. Some early methods~\cite{vo2022noc, narasimhan2018straight, wu2016ask, li2017incorporating, wang2017fvqa, chen2021zero} use existing knowledge bases such as WordNet~\cite{miller1995wordnet}, Wikidata~\cite{vrandevcic2014wikidata}, or collected knowledge as external knowledge sources. By searching for relevant facts in these bases or exploring adjacent nodes in knowledge graphs, these methods extract external knowledge (typically through textual descriptions) to provide a foundational understanding of the unfamiliar visual elements in novel samples. However, these methods often suffer from the problem of complex and inflexible knowledge extraction processes and limited richness of available knowledge content. This hinders these methods from effectively processing novel visual samples, as the complexity of knowledge extraction and the limitations of knowledge content and modality may lead to insufficient or inaccurate contextual information.

\subsection{Task Diversity}
The variety of vision-language tasks leads to great differences in the reasoning process and input-output workflow between different tasks. For example, the image editing task focuses on modifying specific visual elements within an image, where the reasoning involves identifying specific visual elements for alteration based on input instructions and generating precise image modifications as the output. In contrast, the visual entailment task focuses on reasoning about the logical relationships between an image and a textual hypothesis, with outputs restricted to predefined labels such as ``entailment'', ``contradictory'', or ``neutral''.

Constrained by this diversity, most supervised methods specify a single model to handle a specific vision-language task. When handling multiple tasks, methods usually follow the paradigm of training multiple target models. Each model has its own set of parameters that are updated during a specific training process. However, this paradigm imposes additional complexities in practical applications and lacks interpretability.

As a potential solution to this problem, multi-task learning~\cite{Hu_2021_ICCV,nguyen2019multi,lu202012} typically involves an encoder shared between various tasks, such as image captioning, visual question answering, and visual entailment, and multiple task-specific decoders. This architecture allows the model to benefit from shared visual features and visual-textual patterns learned from one task to improve the performance of other tasks. However, successful multi-task learning for vision-language tasks requires carefully designed model architectures and balanced parameter update strategies, which can be challenging in practice. Moreover, when adapting a multi-task model to a new vision-language task, the visual-textual patterns learned by the model and stored in its parameters may not be optimal for the new task, and continuously updating these parameters may suffer from the catastrophic forgetting issue.

\section{Recent Advances in Pre-trained Models}
    
   \subsection{Large Language Models}
    
   Through the new paradigm of ``pre-train, prompt, and predict’’, LLMs have demonstrated remarkable capabilities that scale with the increasing amount of training data and model parameters. LLMs trained in an unsupervised manner are able to perform a variety of tasks through appropriate prompt engineering.
    
    Specifically, LLMs can handle unseen samples via few-shot prompting~\cite{brown2020language} at inference time, where the task pattern is learned from a few input-output examples provided in the context of user input. This technique involves providing a small number of input-output examples within the context of user input, which empowers the model to learn task patterns and apply them to new instances. This flexibility is a significant advantage, as it enables LLMs to adapt to a variety of tasks without extensive retraining. 
    
    In addition to few-shot prompting, instruction fine-tuning~\cite{wei2021finetuned,zhang2023instruction} also plays a crucial role in improving the performance of LLMs. This technique enhances the ability of LLMs to follow instructions, enabling them to perform new tasks based on the requirements of the input instructions. By training models on a diverse set of instructions, instruction fine-tuning equips LLMs with the capability to perform a wide range of tasks while being closely aligned with user intent, thus improving their effectiveness in real-world applications.
    
    A significant breakthrough in the development of LLMs is the technique of reinforcement learning with human feedback (RLHF)~\cite{ouyang2022training,christiano2017deep}. This technique is essential for aligning LLMs with human preferences, ensuring that LLMs are helpful, honest, and harmless. For example, InstructGPT~\cite{ouyang2022training} employs an effective RLHF-tuning approach that enables LLMs to follow expected instructions, which integrates human feedback into the training process. RLHF has been widely adopted in existing LLMs, such as ChatGPT, Claude, and Bard, significantly improving their overall utility and safety in various applications.
    
    \subsection{Vision-Language Models}
    
    Inspired by the success of LLMs in natural language processing, VLMs have gained significant attention in the field of multi-modal learning. Unlike LLMs, which are limited to processing a single data modality, VLMs excel in understanding and generating both visual content and text.
    
    As a representative contrastive VLM, CLIP breaks the traditional supervised learning strategy of vision-language models that requires labeled datasets. In contrast, CLIP directly learns from raw text about images via a simple pre-training task that predicts which caption matches which image. It encodes the input images and text with an image encoder and a text encoder, respectively, and calculates the similarities of the encoded image and text embeddings. Through contrastive pre-training on the WebImageText dataset that contains 400M image-text pairs collected from the internet, CLIP learns a shared semantic space for visual and textual modalities. This insight inspires subsequent research on discriminative VLMs, such as ALIGN~\cite{jia2021scaling}, ALBEF~\cite{li2021align}, Llip~\cite{lavoie2024modeling}, VideoCLIP~\cite{xu2021videoclip}, ActionCLIP~\cite{wang2021actionclip}, etc. 
 
    In addition to the above contrastive VLMs, generative VLMs also play an important role in vision-language tasks. From the perspective of pre-training techniques, BLIP~\cite{li2022blip} introduces a new model architecture for effective multi-task pre-training, called Multimodal mixture of Encoder-Decoder (MED), which contains an image-grounded text encoder and an image-grounded decoder in addition to unimodal encoders.  Thanks to MED, BLIP is pre-trained with three vision-language objectives: image-text contrastive learning, image-text matching, and image-conditioned language modeling. Since then, many new techniques for VLMs have emerged and resulted in various powerful VLMs, such as Q-Former proposed in BLIP2~\cite{li2023blip}, the diffusion process in Stable Diffusion~\cite{rombach2022high}, factually augmented RLHF in LLaVA-RLHF~\cite{sun2023aligning}, Direct Preference Optimization (DPO) in Diffusion-DPO~\cite{yang2024using}, etc.
 
    From the perspective of pre-training data, existing vision-language datasets are suitable choices for VLM pre-training, such as COCO~\cite{lin2014microsoft}, Visual Genome~\cite{krishna2017visual}, Conceptual Captions~\cite{sharma2018conceptual}, and Webvid-2M~\cite{bain2021frozen}. Apart from these datasets, many novel datasets have been constructed to improve the performance of VLMs. Specifically, Liu et al.~\cite{liu2024visual} present a data reformation pipeline to convert image-text pairs into an appropriate instruction-following format using ChatGPT/GPT-4. LLaVA uses 158,000 collected language-image instruction-following samples as data, leading the way in visual instruction tuning for VLMs and showing satisfying instruction-following performance. Similarly, Li et al.~\cite{li2023videochat} build a novel video-centric multimodal instruction-tuning dataset. This dataset comprises thousands of videos associated with detailed descriptions and conversations, which offers a valuable resource for training VideoChat. Subsequently, Li et al.~\cite{li2023mimic} present the MIMIC-IT dataset, which includes multi-modal in-context information, such as multiple instruction-response pairs and multiple images or videos. Training on this dataset enables Otter~\cite{li2023otter} to exhibit remarkable proficiency in in-context learning. 

    \subsection{Benchmarks}
    
    Benchmarks are essential in evaluating the capabilities of VLMs and providing a reliable data source for comparing the performance of different VLMs. They enable researchers to systematically evaluate and validate models, ensuring that advancements in VLMs are both measurable and meaningful in various tasks.
    
    Standard vision-language benchmarks cover evaluations on a variety of common vision-language tasks, such as the COCO~\cite{lin2014microsoft} and Nocaps~\cite{agrawal2019nocaps} benchmarks for image captioning, the OK-VQA~\cite{marino2019ok} and ViZWiZ VQA~\cite{gurari2018vizwiz} benchmarks for visual question answering, the RefCOCO~\cite{yu2016modeling} and RefCOCO+~\cite{yu2016modeling} benchmarks for refer expression comprehension, and the SEED-Bench~\cite{li2023seed} and MMBench~\cite{liu2025mmbench} benchmarks for instruction-following, etc.
    
    In addition to these benchmarks that evaluate the fundamental performance of VLMs, some novel benchmarks are designed to explore the specific capabilities and limitations of VLMs. By using these benchmarks to challenge VLMs, researchers can gain deeper insights into the performance of VLMs and identify areas for further improvement.
 
    For CLIP-like contrastive VLMs, the most commonly used evaluation task in benchmarks is image-text retrieval, where the design of samples varies. Specifically, CountBench~\cite{paiss2023teaching} is an image-text counting benchmark for evaluating VLMs’ understanding of object counting. The MMVP benchmark~\cite{tong2024eyes} exposes nine basic visual patterns, which is specifically designed to inquire about differences in CLIP-blind pairs and evaluate the visual abilities of VLMs using straightforward questions. The Compun benchmark~\cite{kumar2024vision} aims to evaluate the ability of VLMs to interpret compound nouns. Given a text prompt with a compound noun, the task for VLMs is to select the correct image that shows the compound noun among a pair of distractor images that show the constituent nouns that make up the compound noun. Similar to Compun, the Cola benchmark~\cite{ray2024cola} evaluates the compositional reasoning ability of VLMs, where the VLMs are asked to retrieve images with the correct configuration of attributes and objects, and avoid selecting a distractor image with the same objects and attributes but in the wrong configuration. Moreover, CREPE~\cite{ma2023crepe}, ARO~\cite{yuksekgonul2022and}, SUGARCREPE~\cite{hsieh2024sugarcrepe} are also compositionality benchmarks that explore the ability of VLMs to identify relevant captions for a given image in a set of compositional distractors.
 
    For BLIP-like generative VLMs, the corresponding benchmarks mainly focus on investigating the hallucinations of VLMs in different situations. For example, the collected samples in HALLUSIONBENCH~\cite{guan2024hallusionbench} include unedited images such as charts, maps, and posters, as well as hand-crafted or edited images, covering topics such as math, counting, culture, cartoons, sports, geography, sports, and more. Apart from benchmarks focusing on hallucinations, more and more interesting benchmarks have been developed. For example, MultipanelVQA~\cite{fan2024muffin} tests the multi-panel visual reasoning abilities of VLMs, while SOK-Bench~\cite{wang2024sok} evaluates the commonsense reasoning abilities of VLMs in open-world videos. The WHOOPS benchmark~\cite{bitton2023breaking} evaluates the counter-intuitive reasoning abilities, and C-VQA~\cite{zhang2024if} examines the counterfactual reasoning abilities.

\section{Solutions to challenges}
\label{sec:method}
 
\begin{figure*}[!t]
\centering
\includegraphics[width=0.80\textwidth]{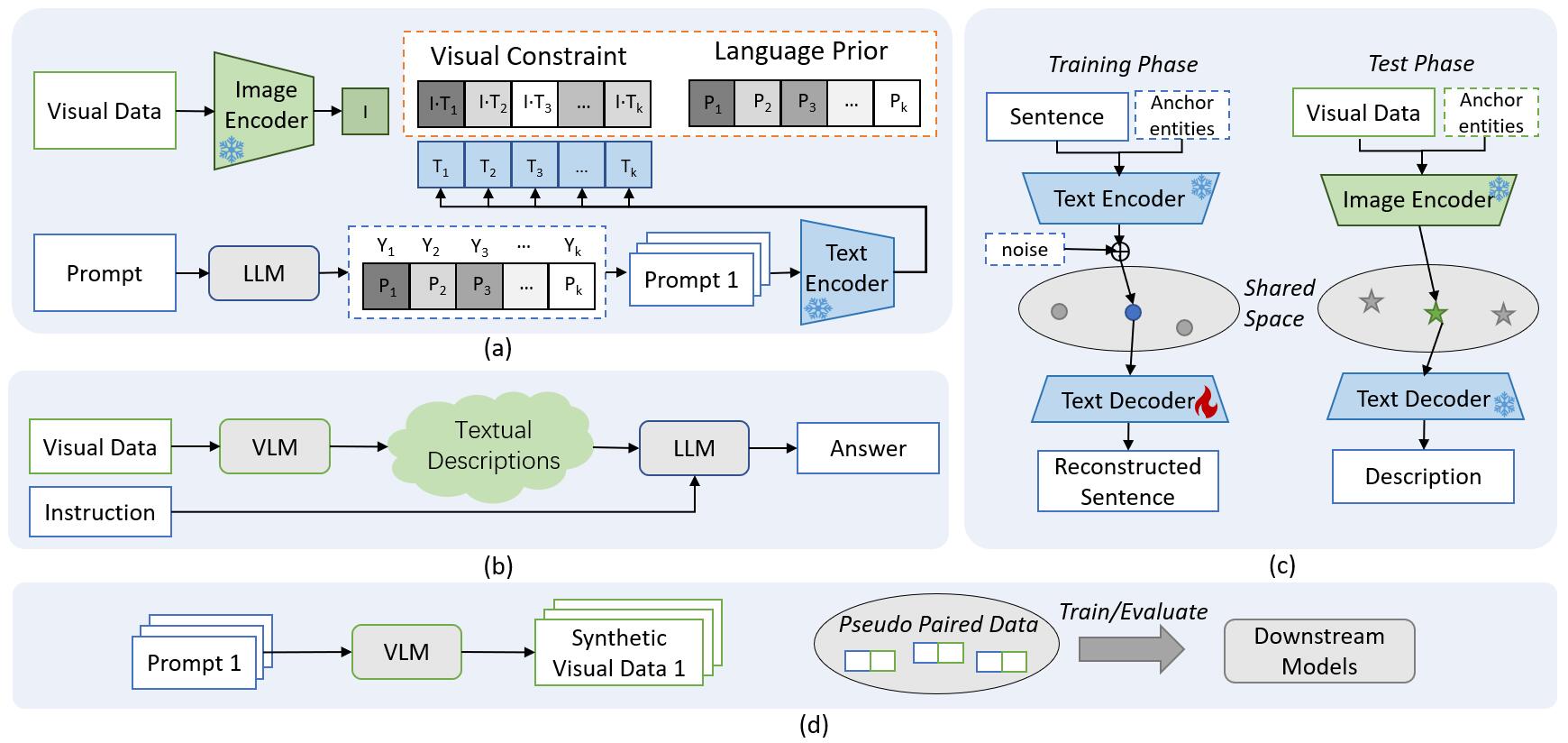}
\caption{Paradigms for addressing the data scarcity challenge in vision-language tasks with the help of pre-trained models. (a) shows the paradigm of integrating an LLM with CLIP to perform direct inference on the test sample.
 (b) shows the paradigm of converting visual information into texts by a VLM to perform direct inference on the test sample.  (c) shows the paradigm of learning from unlabeled uni-modal data.  (d) shows the paradigm of using a VLM to generate pseudo paired data for training or evaluation.}
\label{fig1}
\end{figure*}

In this section, we introduce existing methods for vision-language tasks, which integrate pre-trained models. To provide a structured overview, we categorize these methods according to the challenges they address, which is consistent with the four challenges outlined in Section~\ref{sec:prob}. 
Table~\ref{tab:method} summarizes the target challenges, the followed paradigms, and the overall introductions of the corresponding methods covered by this survey.

\subsection{Solutions to Data Scarcity}

Thanks to the image-text common space learned by VLMs, the language priors captured by LLMs from large-scale corpus,  and the multi-modal transfer capabilities of generative pre-trained models, a series of methods~\cite{tewel2022zerocap,su2022language,zeng2023conzic,zeng2024meacap,wang2023text,tewel2022zero,jo2023zero,song2022clip,yu2022towards,yang2022empirical,tiong2022plug,wang2022language,hanu2023language,bhattacharya2023video,pan2023retrieving,nukrai2022text,gu2022can,li2023decap,wang2023association,wang2022zero,fei2023transferable,kim2023language,wang2022clip,ma2024image,liu2024improving,yang2023freemask,zhou2022towards,bitton2023breaking} have recently emerged to address the data scarcity challenge for the vision-language tasks. 
These methods eliminate the reliance on manually annotated data or require only a small number of paired samples to support inference, and can be broadly categorized into three categories:(1)
Some methods~\cite{tewel2022zerocap,su2022language,zeng2023conzic,zeng2024meacap,wang2023text,tewel2022zero,jo2023zero,song2022clip,yu2022towards,yang2022empirical,tiong2022plug,wang2022language,hanu2023language,bhattacharya2023video,pan2023retrieving} directly perform inference on a single test sample, as shown in Fig.~\ref{fig1}(a) and (b); (2) some methods~\cite{nukrai2022text,gu2022can,li2023decap,wang2023association,wang2022zero,fei2023transferable,kim2023language,wang2022clip} use unlabeled data from the target modality for training and then replace  input features with source modality features from the common space during testing, as shown in Fig.~\ref{fig1}(c); (3) Other methods~\cite{ma2024image,liu2024improving,yang2023freemask,zhou2022towards,bitton2023breaking} employ generative pre-trained models to automatically generate pseudo annotations for training or evaluation purposes, as illustrated in Fig.~\ref{fig1}(d). 
In the following sections, we will review these methods in more details.

\subsubsection{Direct inference on test samples}
\label{sec:inf-test}

In the absence of available annotated data, a series of methods~\cite{tewel2022zerocap,su2022language,zeng2023conzic,zeng2024meacap,wang2023text,tewel2022zero,jo2023zero,song2022clip,yu2022towards,yang2022empirical,tiong2022plug,wang2022language,hanu2023language,bhattacharya2023video,pan2023retrieving} integrate an LLM with CLIP to perform  direct inference on test samples without paired training data. 
As shown in Fig.~\ref{fig1}(a), these methods employ an LLM to provide language priors  during pre-training, while using CLIP to calculate image-text similarity, thereby introducing visual constraints into the test process.

Specifically, to achieve zero-shot image captioning, Tewel et al.~\cite{tewel2022zerocap} propose ZeroCap, which uses GPT-2~\cite{radford2019language} to progressively generate image captions. 
Considering that GPT-2 cannot directly perceive images, this method encodes the captions generated by GPT-2 and the images into the CLIP common space and minimizes the cosine distance between these embeddings to adjust the context cache of GPT-2. 
This operation ensures that the caption matches the visual content. Meanwhile, to maintain grammatical accuracy, ZeroCap uses cross-entropy loss to constrain the generated content to be consistent with the language-prior knowledge in GPT-2. 
Similarly, Su et al.~\cite{su2022language} propose a new decoding strategy that incorporates CLIP similarity scores into the decoding probability distributions of GPT-2. This strategy ensures the obtained token probability distributions take into account the similarity between image and text. 
Zeng et al.~\cite{zeng2023conzic} enhance ZeroCap~\cite{tewel2022zerocap} by introducing Gibbs-BERT, which adopts sampling-based search instead of the auto-regressive generation used in ZeroCap. In addition, they incorporate a task-specific discriminator to identify captions that align with the user control signal. 
To generate captions rich in world knowledge, Zeng et al.~\cite{zeng2024meacap} design a retrieve-then-filter module to obtain key concepts relevant to the image. Using these concepts, they employ a pre-trained keywords-to-sentence LLM called CBART~\cite{he2021parallel} to generate corresponding image captions under the guidance of visual relevance fusion scores. 
For visual storytelling, Wang et al.~\cite{wang2023text} formulate this task as a visual-conditioned generation problem, where a visual condition planner aggregates the visual conditions computed by CLIP into GPT-2.

Later, this paradigm has been extended into the video captioning task. For instance, Tewel et al.~\cite{tewel2022zero} modify the token-level optimization in ZeroCap to sentence-level optimization, as token-level optimization tends to force each token to describe all frames in the video. This method introduces a cross-entropy loss similar to ZeroCap, and uses total frame-caption matching scores from CLIP to update the pseudo tokens in the GPT-2 prompt, instead of adjusting the GPT-2 context cache. 
For dense video captioning, Jo et al.~\cite{jo2023zero} use GPT-2 and CLIP to localize and describe events in videos. In their method, a soft moment mask is introduced to represent video temporal segments. This method jointly optimizes the soft moment mask and the textual prefix context, aiming to accurately align the generated text with the corresponding events in the video.

Similarly, in the visual question answering (VQA) task, Song et al.~\cite{song2022clip} convert questions into masked templates and then fill these templates using inherent commonsense knowledge from T5~\cite{raffel2020exploring} to obtain multiple candidate answers. The candidate answer with the highest CLIP similarity value to the image is selected as the final answer. 
Yu et al.~\cite{yu2022towards} use StyleGAN~\cite{karras2019style} as a generator for counterfactual image editing, during which CLIP is used to ensure the alignment between generated images and the target counterfactual text.

In addition to combining the language priors from an LLM and the visual constraints from CLIP, another feasible idea is to use a VLM to convert visual information into texts and then process these texts by an LLM to complete the task, as shown in Fig.~\ref{fig1}(b).

Yang et al.~\cite{yang2022empirical} verbalize visual information in an image using its caption. 
They then prompt  GPT-3~\cite{brown2020language} to address the target VQA task by inputting the concatenation of the image caption and the question, with the caption as contextual information. 
This allows  generating answers to questions in an open-ended text generation manner, applicable to both zero-shot and few-shot settings. 
Similarly, Tiong et al.~\cite{tiong2022plug} propose a method that generates captions for the top-k image patches most relevant to the question. A question-answering module is then designed to generate an answer based on the question and the generated captions.  

Since videos contain more information than images, video-oriented methods~\cite{wang2022language,hanu2023language,bhattacharya2023video,pan2023retrieving} employ more diverse tools to describe videos rather than just captions. 
Wang et al.~\cite{wang2022language} extract video information at both the visual-token level and the frame level to generate comprehensive video descriptions. In addition,  Wang et al.~\cite{wang2022language} and Hanu et al.~\cite{hanu2023language} both use automatic speech recognition to encode audio information in videos. Bhattacharya et al.~\cite{bhattacharya2023video} further enhance video descriptions by extracting subtitles from video frames using optical character recognition. For a given video, Pan et al.~\cite{pan2023retrieving} retrieve a set of semantically similar texts from an external corpus. By using these retrieved texts along with the question, an LLM can then generate the final answer.

\begin{table}[t]
  \centering
  \caption{Performance comparison of different image captioning methods proposed to address the data scarcity challenge.}
  \resizebox{0.98\columnwidth}{!}{
    \begin{tabular}{l|cccc|cccc}
\toprule
\multicolumn{1}{c|}{\multirow{2}[3]{*}{Methods}} & \multicolumn{4}{c|}{COCO}   & \multicolumn{4}{c}{Flickr30k} \\
\cmidrule{2-9}          & \multicolumn{1}{c}{B@4} & \multicolumn{1}{c}{M} & \multicolumn{1}{c}{C} & \multicolumn{1}{l|}{S} & \multicolumn{1}{c}{B@4} & \multicolumn{1}{c}{M} & \multicolumn{1}{c}{C} & \multicolumn{1}{c}{S} \\
    \midrule

          & \multicolumn{8}{c}{Supervised training} \\
    \midrule
    \multicolumn{1}{c|}{BUTD~\cite{anderson2018bottom}} & \multicolumn{1}{c}{36.2} & \multicolumn{1}{c}{27.0} & \multicolumn{1}{c}{113.5} & \multicolumn{1}{c}{20.3} & \multicolumn{1}{c}{27.3} & \multicolumn{1}{c}{21.7} & \multicolumn{1}{c}{56.5} & \multicolumn{1}{c}{16.0} \\
    \multicolumn{1}{c|}{ClipCap~\cite{mokady2021clipcap}} & \multicolumn{1}{c}{33.5} & \multicolumn{1}{c}{27.5} & \multicolumn{1}{c}{113.1} & \multicolumn{1}{c}{23.2} & \multicolumn{1}{c}{21.7} & \multicolumn{1}{c}{22.1} & \multicolumn{1}{c}{53.5} & \multicolumn{1}{c}{-} \\
    \midrule
          & \multicolumn{8}{c}{Direct inference on test samples} \\
    \midrule
    ZeroCap\cite{tewel2022zerocap} & 2.6   & 11.5  & 14.6  & 5.5   & \multicolumn{1}{c}{-} & \multicolumn{1}{c}{-} & \multicolumn{1}{c}{-} & \multicolumn{1}{c}{-} \\
    MAGIC\cite{su2022language} & 12.9  & 17.4  & 49.3  & 11.3  & 6.4   & 13.1  & 20.4  & 7.1 \\
    ConZIC\cite{zeng2023conzic} & 1.3   & 11.2  & 13.3  & 5.0     & \multicolumn{1}{c}{-} & \multicolumn{1}{c}{-} & \multicolumn{1}{c}{-} & \multicolumn{1}{c}{-} \\
    MeaCap\cite{zeng2024meacap} & 7.1   & 16.6  & 42.5  & 11.8  & 7.2   & 17.8  & 36.5  & 13.1 \\
    CEPT\cite{tewel2022zero} & 2.2   & 12.7  & 17.2  & 7.3   & \multicolumn{1}{c}{-} & \multicolumn{1}{c}{-} & \multicolumn{1}{c}{-} & \multicolumn{1}{c}{-} \\
    \midrule
          & \multicolumn{8}{c}{Text-only training} \\
    \midrule
    CapDec\cite{nukrai2022text} & 26.4  & 25.1  & 91.8  & 11.9  & 17.7  & 20.0    & 39.1  & 9.9 \\
    CLOSE\cite{gu2022can} & 28.6  & 25.2  & 95.4  & 18.1  & \multicolumn{1}{c}{-} & \multicolumn{1}{c}{-} & \multicolumn{1}{c}{-} & \multicolumn{1}{c}{-} \\
    DeCap\cite{li2023decap} & 24.7  & 25.0    & 91.2  & 18.7  & 21.2  & 21.8  & 56.7  & 15.2 \\
    Knight\cite{wang2023association} & 27.8  & 26.4  & 98.9  & 19.6  & 22.6  & 24.0    & 56.3  & 16.3 \\
    CLMs\cite{wang2022zero} & 15.0    & 18.7  & 55.7  & 10.9  & 16.8  & 16.2  & 22.5  & 9.8 \\
    ViECap\cite{fei2023transferable} & 27.2  & 24.8  & 92.9  & 18.2  & 21.4  & 20.1  & 47.9  & 13.6 \\
    \midrule
          & \multicolumn{8}{c}{Generate pseudo paired data} \\
    \midrule
    ICSD\cite{ma2024image}  & 29.9  & 25.4  & 96.6  & \multicolumn{1}{c|}{-} & 25.2  & 20.6  & 54.3  & \multicolumn{1}{c}{-} \\
    SynTIC\cite{liu2024improving} & 29.9  & 25.8  & 101.1 & 19.3  & 22.3  & 22.4  & 56.6  & 16.6 \\
    \bottomrule
    \end{tabular}}%
  \label{tab:data}%
\end{table}%

\subsubsection{Learning from unlabeled uni-modal data}
\label{sec:uni-modal}

CLIP learns to align image and text embeddings during pre-training. Inspired by this insight, an intuitive idea to address the data scarcity challenge is to learn from unlabeled uni-modal data as an approximation of supervised training on paired vision-language data. 
When only text data is available, the paired visual data required for training can be replaced by the CLIP embeddings of  text data. These text embeddings together with the corresponding text data form paired data for supervised training on the target task. During the test phase, the text embeddings are replaced by the visual embeddings encoded by CLIP. And vice versa if only visual data is available. Fig.~\ref{fig1}(c) shows this idea of learning from unlabeled uni-modal data.
However, the domain gap between the CLIP embeddings of the two modalities presents a challenge to the success of this ideal replacement. 

Nukrai et al.~\cite{nukrai2022text} make the first attempt to reduce this gap. They hypothesize that in CLIP's visual-textual common space, the visual embedding lies within a small sphere around the corresponding text embedding. By injecting Gaussian noise into the input text embedding during training, all embeddings within this sphere are mapped to the same caption. Based on this assumption, the corresponding image embedding should also fall into this sphere and can be decoded to the correct caption. Similarly, Gu et al.~\cite{gu2022can} find that adding Gaussian noise is effective and also explore using mean shift, linear adapters, and structured noise as additional methods to bridge the gap. 
Inspired by these findings, many efforts have made to solve the gap problem. 
 Li et al.~\cite{li2023decap} and Wang et al.~\cite{wang2023association} both replace the image embedding with a weighted combination of similar text embeddings during the test phase. Moreover, Wang et al.~\cite{wang2023association} replace the input text embeddings with similar weighted combinations during training to further narrow the training-test gap. Some methods use entities in sentences and objects in images as anchors~\cite{wang2022zero} or additional prompts~\cite{fei2023transferable} for the decoder to make the training input and test input more similar.

In addition to the scenarios with only text data available, several other methods ~\cite{kim2023language,wang2022clip,esser2021taming}  focus on using solely visual data to achieve a text-free training process.  
Kim et al.~\cite{kim2023language} propose a zero-shot video localization method that selects a frame and encodes it using CLIP as a pseudo-language query representation. The frame is selected from an event proposal generated based on the visual similarity of frames. During the test phase, the pseudo-language query representation is replaced by the real language query embeddings from CLIP. Similarly, CLIP-GEN~\cite{wang2022clip} only requires  a set of unlabeled images to train a text-to-image generator. Specifically, it trains a Transformer to convert image CLIP embeddings into discrete image tokens in the VQGAN~\cite{esser2021taming} codebook space. After training, the Transformer can generate coherent image tokens from the input text CLIP embeddings, which can then be decoded into images by VQGAN.

\subsubsection{Generating pseudo paired data}

When no paired data is  available for training, another feasible solution is to leverage the multi-modal conversion capabilities of pre-trained models to automatically generate pseudo annotations for training or evaluation, as illustrated in Fig.~\ref{fig1}(d). 

Using tools like Stable Diffusion~\cite{rombach2022high}, synthetic images can be generated to create pseudo-image-sentence pairs for training an image captioning model. 
Specifically, Ma et al.~\cite{ma2024image} propose a method to generate multi-context synthetic images by summarizing captions describing the same image from different perspectives. Liu et al.~\cite{liu2024improving} further refine pseudo image features to make them closer to natural image features. Yang et al.~\cite{yang2023freemask} synthesize a variety of new images based on semantic masks in the target dataset, forming new training pairs with the synthetic images and conditional masks for semantic segmentation. Instead of generating data, Zhou et al.~\cite{zhou2022towards} use pseudo-text features derived from images to learn the text-to-image generation, eliminating the need for a general image captioning model.   
This paradigm is also applicable when the expected data is difficult to collect from the real world, such as counterfactual images~\cite{bitton2023breaking} containing co-occurring elements that violate commonsense knowledge.  

In summary, by integrating powerful pre-trained models, the aforementioned methods better bridge the gap between visual and textual modalities than methods before the pre-trained model era. Specifically, previous methods typically connect two modalities from scratch solely based on their shared objects or relationships. Their core components or ideas include applying object-based rewards~\cite{feng2019unsupervised}, treating objects as anchors to learn a shared manifold~\cite{laina2019towards}, representing data in both modalities by objects~\cite{guo2020recurrent}, constructing pseudo-paired data based on object relationships~\cite{qi2024relational}, etc. Compared to the extensive knowledge encapsulated in pre-trained models, the information available in objects or relationships is limited, which hinders previous methods from effectively bridging the modality gap and often results in noisy outputs.

For the above three categories of methods to solve the challenge of data scarcity, we take the visual captioning task with the largest number of related works as an example. We summarize the performance of these methods~\cite{tewel2022zerocap,su2022language,zeng2023conzic,zeng2024meacap,tewel2022zero,nukrai2022text,gu2022can,li2023decap,wang2023association,wang2022zero,fei2023transferable,ma2024image,liu2024improving} and compare them with two classic fully supervised methods~\cite{anderson2018bottom,mokady2021clipcap}. As representative benchmarks for the visual captioning task, COCO~\cite{lin2014microsoft} and Flickr30K~\cite{plummer2015flickr30k} are also widely used to evaluate methods addressing data scarcity in this task. The COCO dataset consists of 123,287 images, including 5,000 images for validation and 5,000 images for testing. Each image in COCO has about five crowdsourced captions. The Flickr30K dataset contains 31,783 images, including 1,000 images for validation, and 1,000 images for testing. Each image in Flickr30K is annotated with five crowdsourced captions. BLEU4~\cite{papineni2002bleu}, METEOR~\cite{banerjee2005meteor}, CIDEr~\cite{vedantam2015cider}, and SPICE~\cite{anderson2016spice} are commonly used evaluation metrics for evaluating the quality of the generated text. BLEU4 measures precision by calculating the overlap of four-grams between the generated captions and the reference captions. METEOR considers precision and recall by aligning the generated captions with the reference captions based on stemming and synonymy. It provides a more human-like evaluation by accounting for variations in wording. As a specialized metric for visual captioning, CIDEr highlights important words by weighting n-grams according to their frequency in reference captions. SPICE evaluates the semantic content of generated captions by focusing on the mentioned objects, attributes, and relationships. Table~\ref{tab:data} shows the comparison of methods on the COCO and Flickr30K datasets using the aforementioned four evaluation metrics. 

As shown in Table~\ref{tab:data}, the methods that generate pseudo paired-data achieve the best performance, comparable to the fully supervised methods. The second-best are the methods that use  uni-modal data for training, while the methods that directly perform inference on test samples perform slightly worse. 
The direct inference methods are superior in their training-free characteristic and are easy to apply in practice. Such methods only need a  test visual sample as input data and do not require additional unimodal training data such as image datasets or sentence corpus. However, due to the absence of sentence corpus as a reference for language style, these methods have difficulty in generating content consistent with the ground-truth caption language style, resulting in lower scores in n-gram-based evaluation metrics. In addition, some of them use CLIP to calculate image-text similarity and sequentially select words to complete the masked sentence based on the calculated similarity scores. Such methods may encounter the risk of concept association bias caused by CLIP acting as a bag-of-words model, which will be discussed in Section~\ref{sec:bag}. 
The idea of learning from unlabeled uni-modal data may suffer from the discrepancy between training and inference, where text embeddings are used for training while visual embeddings are for inference. Fortunately, methods based on this idea keep exploring ways to narrow the modality gap between the visual and textual embeddings of CLIP, which not only brings consistent performance improvements but also provides valuable insights for subsequent studies in vision-language tasks. 
For methods that generate pseudo paired data, the strong multi-modal conversion capabilities of pre-trained models enable them to effectively establish the connection between visual and textual modalities. Their main drawback lies in the error accumulation during the construction of pseudo paired data and the subsequent training process. Moreover, these methods need to develop strategies to enrich the synthetic training data, so that the trained model can be sufficiently robust during inference. On the other hand, the cost of constructing pseudo paired data also needs to be considered when applying these methods in practice, especially the expensive visual data generation.

\begin{figure*}[!t]
\centering
\includegraphics[width=0.8\linewidth]{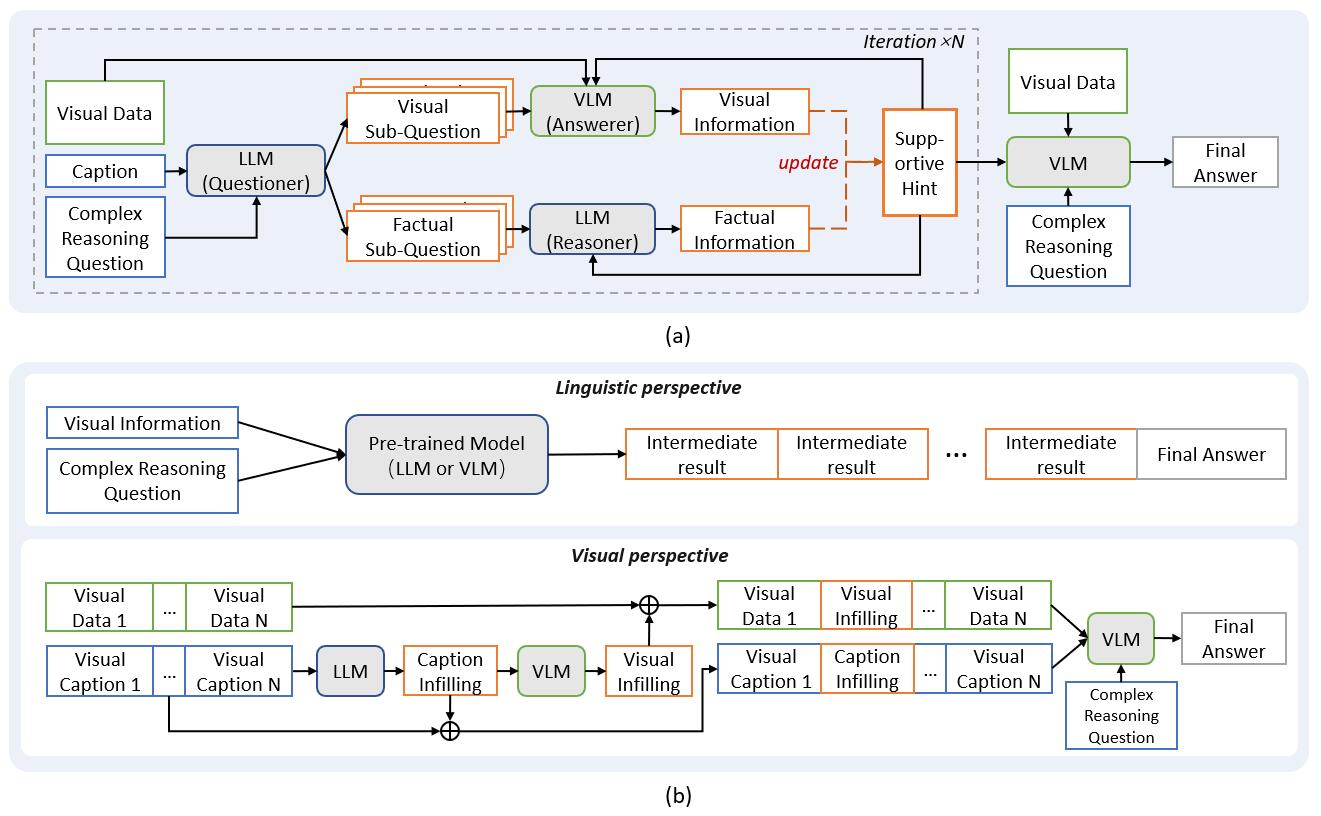}
\caption{Paradigms of using pre-trained models to conquer the challenge of escalating reasoning complexity in vision-language tasks. (a) shows the basic idea of the divide-and-conquer solution. (b) shows the chain-of-thought solution, which includes two different pipelines for linguistic and visual perspectives.}
\label{fig2}
\end{figure*}

\subsection{Solutions to Escalating Reasoning Complexity}

As the reasoning complexity in vision-language tasks, such as visual question answering, visual entailment, and visual commonsense reasoning, continues to increase, it becomes increasingly challenging  for models to accurately predict answers through a single reasoning step. 
To address this challenge,  divide-and-conquer paradigms~\cite{chen2023see,zhou2023vicor,wang2023domino,you2023idealgpt,yang2023good,qi2023the,wang-etal-2023-filling,rajabzadeh2023multimodal} and  chain-of-thought (CoT) paradigms~\cite{lu2022learn,zhang2023multimodal,wang2024t,mondal2024kam,chen2023measuring,zhu2023efficient,mitra2023compositional,zhang2024cocot,meng2023chain,rose2023visual,himakunthala2023let} have been proposed to perform multi-step reasoning by respectively decomposing the original question and the prediction process, thus enhancing the accuracy of answers to complex questions, as shown in Fig.~\ref{fig2}.

\subsubsection{Divide-and-conquer}
\label{sec:divide}

Thanks to the synergy between LLMs and VLMs, the main question in a complex visual-language reasoning task can be decomposed into a series of  sub-questions about visual details and  sub-questions about factual knowledge related to the main question. Since the sub-questions are easier  for models to answer than the original question and the  information answered helps reasoning, the divide-and-conquer paradigm reduces the reasoning complexity by answering the divided sub-questions and integrating their answers to predict the final answer.  This divide-and-conquer paradigm is shown in Fig.~\ref{fig2} (a).

A series of studies~\cite{chen2023see,zhou2023vicor,wang2023domino,you2023idealgpt,yang2023good} focus on exploring more visual information to help models reason about a given task. Some of these studies~\cite{chen2023see,zhou2023vicor,wang2023domino} refine the original practice of using general visual captions to represent visual content and feed it into an LLM for reasoning. For instance, Chen et al.~\cite{chen2023see} design a ``see, think and confirm'' procedure to extract detailed object information from images. However, this method uses a fixed perception process across different questions and images. In contrast, Zhou et al.~\cite{zhou2023vicor} develop a novel framework to automatically unveil crucial visual details related to the reasoning question, which is achieved by querying a VLM with sub-questions generated by an LLM. 
This paradigm is also applicable to handling information-dense images~\cite{wang2023domino}, such as charts or plots. 

Furthermore, to ensure that the LLM gathers sufficient information, You et al.~\cite{you2023idealgpt} use ChatGPT as the selected LLM to iteratively decompose the reasoning task. The process continues until the reasoner (\textit{e.g.} LLM) feels confident that it can solve the original problem based on the collected information. 
Similarly, Yang et al.~\cite{yang2023good} use the strong language priors of ChatGPT to generate detailed pre-questions based on the original question. These pre-questions are then directly concatenated  with the original question to form the input prompt for VLM reasoning. 

In addition to visual information, many methods~\cite{qi2023the,wang-etal-2023-filling,rajabzadeh2023multimodal} also emphasize acquiring factual knowledge related to the question and visual content. 
For example, Qi et al.~\cite{qi2023the} propose a method that recursively decomposes a question into simpler factual and visual sub-questions until they are answerable. The factual sub-questions are answered by GPT-3~\cite{brown2020language} to provide factual knowledge, such as the release date of a movie depicted in an image, which cannot be directly observed from the visual content alone. 
After gathering visual and factual hints, Wang et al.~\cite{wang-etal-2023-filling} introduce a refinement module to filter out irrelevant hints and summarize useful ones. 
Rajabzadeh et al.\cite{rajabzadeh2023multimodal} employ a divide-and-conquer strategy, using tool interactions supplemented by web search to provide additional supporting hints.

\begin{table}[!t]
  \centering
  \caption{Performance comparison of divide-and-conquer methods and chain-of-thought methods in solving the challenge of the escalating reasoning complexity.}
  \resizebox{0.98\columnwidth}{!}{
    \begin{tabular}{c|ccccc}
    \toprule
    Methods & OK-VQA & A-OKVQA & VCR   & SNLI-VE & ScienceQA \\
    \midrule
          & \multicolumn{5}{c}{One-step reasoning} \\
    \midrule
    MiniGPT-4~\cite{zhu2023minigpt} & 37.5  & 58.2  & 40.6  & 35.1  & 47.4 \\
    \midrule
          & \multicolumn{5}{c}{Divide-and-conquer} \\
    \midrule
    IPVR\cite{chen2023see}  & 44.6  & 46.0  & -     & -     & - \\
    FIIG\cite{wang-etal-2023-filling} & 59.3  & 59.8 & -     & -     & - \\
    ViCoR\cite{zhou2023vicor} & -     & 70.9  & 55.4  & -     & - \\
    IdealGPT\cite{you2023idealgpt} & -     & -     & 50.7  & 55.3  &  \\
    Qvix\cite{yang2023good}  & -     & -     & -     & 50.1  & 55.0 \\
    \midrule
          & \multicolumn{5}{c}{Chain-of-thought} \\
    \midrule
    mm-CoT\cite{zhang2023multimodal} & -     & -     & -     & -     & 84.9 \\
    T-SciQ\cite{wang2024t} & -     & -     & -     & -     & 91.8 \\
    KAM-CoT\cite{mondal2024kam} & -     & -     & -     & -     & 92.5 \\
    \bottomrule
    \end{tabular}}%
  \label{tab:reason}%
\end{table}%

\subsubsection{Chain-of-Thought}

The chain-of-thought paradigm~\cite{wei2022chain} decomposes the direct prediction process of an LLM into a series of intermediate reasoning steps. 
Inspired by its success in addressing complex natural language problems, many researchers~\cite{lu2022learn,zhang2023multimodal,wang2024t,mondal2024kam,chen2023measuring,zhu2023efficient,mitra2023compositional,zhang2024cocot,meng2023chain,rose2023visual,himakunthala2023let} have tried to adopt this paradigm to solve complex visual-language reasoning problems.

Early methods explore using LLMs for multi-modal chain-of-thought reasoning~\cite{lu2022learn,zhang2023multimodal,wang2024t,mondal2024kam}, where the LLM generates rationales or step-by-step thought processes as intermediate outputs to improve the accuracy of the final answers. 
The main concern of this approach lies in how to effectively inject visual information into an LLM. 
Lu et al.~\cite{lu2022learn} represent visual content through corresponding captions and use this approach to evaluate several LLM baselines on a newly proposed benchmark ScienceQA. This benchmark includes data examples consisting of multi-modal question-answering information and grounded lectures and explanations. Later, Zhang et al.~\cite{zhang2023multimodal} introduce a two-stage framework to sequentially perform rationale generation and rationale-based answer reasoning by fine-tuning an LLM to accept fused image and language features as input. Given the challenge of collecting a high-quality CoT corpus, Wang et al.~\cite{wang2024t} follow a plan-and-solve prompting process to automatically generate CoT rationales in a zero-shot manner. These rationales serve as teaching signals to fine-tune a student VLM. In addition, Mondal et al.~\cite{mondal2024kam} incorporate external knowledge from a knowledge graph during reasoning to provide supplementary contextual information to  an LLM, thereby  improving reasoning performance.

\begin{figure*}[!t]
\centering
\includegraphics[width=0.90\textwidth]{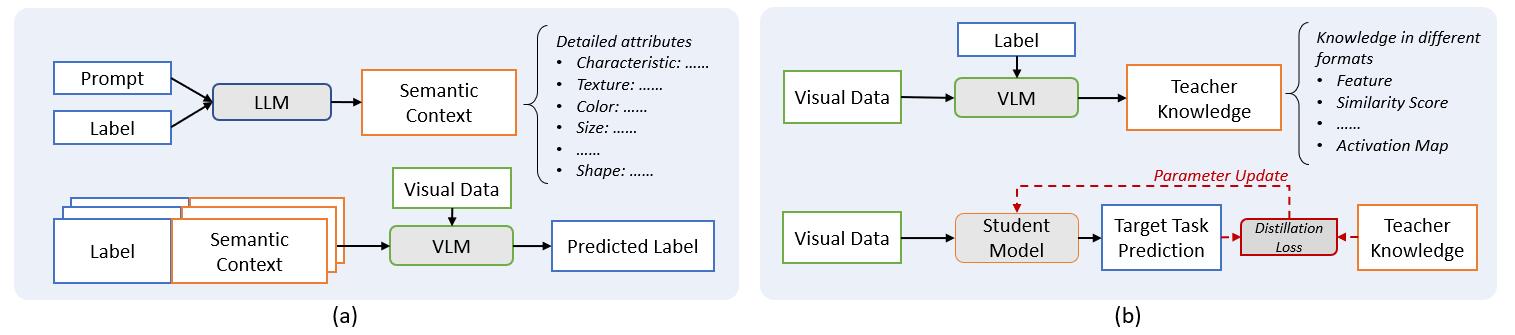}
\caption{Paradigms of using pre-trained models to conquer the generalization challenge to novel samples in vision-language tasks. (a) shows the basic idea of extracting semantic context from an LLM, while (b) shows the basic idea of distilling teacher knowledge from a VLM. }
\label{fig3}
\end{figure*}

Subsequently, the research focus shifts towards exploring the unique chain-of-thought reasoning process for visual-language tasks~\cite{chen2023measuring,zhu2023efficient,mitra2023compositional,zhang2024cocot}, starting with evaluating the chain-of-thought capabilities of various VLMs and then enhancing them. 
Chen et al.~\cite{chen2023measuring} establish the CURE benchmark and evaluate existing VLMs. They observe that  most VLMs have difficulty with CoT reasoning and propose a method for VLMs to learn the CoT reasoning ability from an LLM. This method involves training the VLM to generate rationales under the guidance and feedback of an LLM. For the task of visual document understanding, Zhu et al.~\cite{zhu2023efficient}  integrate outputs from OCR tools, an LLM, and a multimodal-verifier to form teacher rationales. These rationales are then used to train a small student VLM  to predict both rationales and answers for input questions. 
The aforementioned methods still require additional training and support from other models. Mitra et al.~\cite{mitra2023compositional} and Zhang et al.~\cite{zhang2024cocot} propose CCoT and CoCoT, respectively, which directly obtain rationales from the VLM itself through prompting without  extra training. Specifically, CCoT~\cite{mitra2023compositional} regards scene graphs generated by the VLM as intermediate rationales for better compositional reasoning. On the other hand, CoCoT~\cite{zhang2024cocot} prompts the VLM to derive comparison results across multiple input images as intermediate rationales to enhance VLM's reasoning capabilities on tasks involving multi images.  

All the above methods embody the thinking process from a linguistic perspective, as shown at the top of Fig.~\ref{fig2}(b). However, as depicted at the bottom of Fig.~\ref{fig2}(b), another group of methods~\cite{meng2023chain,rose2023visual,himakunthala2023let} seeks to embody this process from a visual perspective by integrating visual augmentation into reasoning. Meng et al.~\cite{meng2023chain} suggest that language-based reasoning is often too complex and abstract. To address this problem, they propose CoI that transforms the conventional textual step-by-step reasoning into a sequence of generated images. This visual augmentation technique is also effective for processing more complex sequential visual data.  
Benefiting from the multi-modal generative capabilities of VLMs,  Rose et al.~\cite{rose2023visual} and Himakunthala et al.~\cite{himakunthala2023let} both aim to bridge the existing logical gaps in sequential data by generating multi-modal fillings. This strategy enriches the temporal and causal information derived from the visual content, thereby improving the reasoning performance. 

Three complex visual reasoning tasks are commonly used to evaluate the reasoning capabilities of methods: knowledge-based visual question answering, visual commonsense reasoning, and visual entailment. For knowledge-based visual question answering, OK-VQA~\cite{marino2019ok}, A-OKVQA~\cite{schwenk2022okvqa}, and ScienceQA~\cite{lu2022learn} are representative benchmarks. OK-VQA is an open-domain dataset comprising 14,055 image-question pairs that span various knowledge domains, including transportation, food, and weather, with each question compared with ten ground-truth answers. This benchmark challenges methods to reason using commonsense and domain-specific knowledge. A-OKVQA extends OK-VQA by scaling up the dataset and task complexity, providing 24,903 samples with justifications for answers. Different from OK-VQA and A-OKVQA, ScienceQA targets at scientific reasoning, covering questions from natural science, social science, and language science. Each question in ScienceQA is paired with detailed annotations, including lectures and explanations of answers. For visual commonsense reasoning, VCR~\cite{zellers2019recognition} consists of 290,000 samples derived from 110,000 unique movie scenes, emphasizing multi-step reasoning on human-centric scenarios. For visual entailment, SNLI-VE~\cite{xie2019visual} is a typical benchmark that requires methods to determine whether an image semantically entails a textual hypothesis. This benchmark contains about 570,000 image-hypothesis-answer triplets. In these benchmarks, the accuracy of generated answers serves as the evaluation metric. We summarize the performance of the aforementioned methods~\cite{chen2023see,wang-etal-2023-filling,zhou2023vicor,you2023idealgpt,yang2023good,zhang2023multimodal,wang2024t,mondal2024kam} addressing the challenge of escalating reasoning complexity and compare them with the one-step reasoning results from the pre-trained model MiniGPT-4~\cite{zhu2023minigpt}. 
Table~\ref{tab:reason} shows the comparison of these methods on three complex visual reasoning tasks, using accuracy as the evaluation metric. 

As shown in Table~\ref{tab:reason}, the divide-and-conquer and chain-of-thought methods outperform  one-step reasoning results of MiniGPT-4 on all datasets. It is worth noting that the  chain-of-thought methods~\cite{zhang2023multimodal,wang2024t,mondal2024kam} also surpass Qvix~\cite{yang2023good} in ScienceQA, since the chain-of-thought methods involve fine-tuning with teacher data from an LLM. 
These comparisons demonstrate that decomposing the original question or the reasoning process is an effective paradigm for solving complex visual reasoning problems. 
Specifically, divide-and-conquer methods enrich the input information that supports pre-trained models to reason complex questions. The enriched input contains detailed visual information and factual knowledge that are useful for reasoning, which is obtained by answering sub-questions divided from the original reasoning question. This paradigm is suitable for addressing complex reasoning questions that require a thorough comprehension of visual information and  invisible background knowledge. However, as the original question is divided into multiple sub-questions, these methods suffer from the risk of error accumulation in the process of collecting the answers of the sub-questions. 
Chain-of-thought methods enhance the reasoning process of pre-trained models by encouraging them to perform step-by-step reasoning and output  thinking processes instead of just providing an isolated answer. This paradigm not only improves reasoning accuracy but also increases the interpretability of integrating pre-trained models to complete vision-language tasks. Moreover, chain-of-thought methods that embody the thinking process from a visual perspective are effective in circumstances where reasoning over complex sequential visual data or step-by-step visual reasoning is required, since these methods are able to use  more complementary visual information for reasoning. Despite the aforementioned strengths, most chain-of-thought methods  involving VLMs require fine-tuning on manually annotated multi-modal chain-of-thought data, and the annotation of such data is expensive and needs expert knowledge.

\begin{table*}[!t]
    \centering
    \begin{minipage}{0.52\textwidth}
        \centering
  \caption{Performance comparison of different open-vocabulary image classification methods proposed to address the challenge of generalization to novel samples. }
   \resizebox{\linewidth}{!}{%
    \begin{tabular}{c|cccccc|c}
    \toprule
    Methods & ImageNet & CUB   & Food101 & Place365 & \makecell[c]{Oxford\\Pets}  & \makecell[c]{Describable\\Texture} & Mean \\
    \midrule
    CLIP\cite{radford2021learning}  & 58.46 & 51.95 & 79.31 & 37.37 & 79.94 & 41.38 & 58.07 \\
    \midrule
    VCD\cite{menon2022visual}   & 62.97 & 52.57 & 83.63 & 39.90  & 83.46 & 44.26 & 61.13 \\
    LCDAtt\cite{yan2023learning} & -     & 64.05 & 81.85 & -     & 85.91 & -     & - \\
    CPHC\cite{ren2024chatgpt}  & 63.88 & 54.18 & 83.02 & 40.73 & 82.69 & 48.19 & 62.12 \\
    \bottomrule
    \end{tabular}}
  \label{tab:class}%
    \end{minipage}\hfill
    \begin{minipage}{0.45\textwidth}
  \centering
  \caption{Performance comparison of different open-vocabulary object detection methods proposed to address the challenge of generalization to novel samples. }
   \resizebox{\linewidth}{!}{%
    \begin{tabular}{c|cc|ccc}
    \toprule
    \multirow{2}[4]{*}{Methods} & \multicolumn{2}{c|}{LVIS} & \multicolumn{3}{c}{COCO} \\
\cmidrule{2-6}          & ${\rm Ap_r}$   & AP    & Novel AP & Base AP & Overall AP \\
    \midrule
    Supervised-RFS~\cite{gupta2019lvis} & 12.3  & 24.3  & -     & -     & - \\
    OVR-CNN~\cite{zareian2021open} & -     & - & 22.8  & 46.0  & 39.9 \\
    \midrule
    ViLD\cite{gu2022open}  & 16.6  & 25.5  & 27.6  & 59.5  & 51.3 \\
    PB-OVD\cite{gao2022open} & -     & -     & 30.8  & 46.1  & 42.1 \\
    FVLM\cite{kuo2023open}  & 18.6  & 24.2  & 28.0    & 43.7  & 39.6 \\
    BARON\cite{wu2023aligning} & 22.6  & 27.6  & 42.7  & 54.9  & 51.7 \\
    OADP~\cite{wang2023object}  & 21.9  & 28.7  & 30.0  & 53.3  & 47.0\\
    \bottomrule
    \end{tabular}}%
  \label{tab:object}%
    \end{minipage}
\end{table*}

\subsection{Solutions to Generalization to Novel Samples}

When a model encounters novel visual samples unseen during the training phase, it cannot make accurate inferences due to the lack of knowledge about the visual appearance and semantic meaning of these samples. Pre-trained models have a large number of parameters are trained extensively on massive datasets, and their parameters contain a wealth of world knowledge. Therefore, from both the visual and semantic perspectives, pre-trained models are suitable choices to provide external knowledge for novel samples. As shown in Fig.~\ref{fig3}, the solutions to the generalization challenge to novel samples can be roughly divided into two categories, including extracting semantic context from an LLM~\cite{menon2022visual,yan2023learning,dai2023exploring,ren2024chatgpt,li2024zero,jia2023generating,yousaf2023videoprompter} and distilling teacher knowledge from a VLM~\cite{gu2022open,wu2023aligning,wang2023object,gao2022open,kuo2023open}.

\subsubsection{Extracting semantic context from an LLM}
\label{sec:LLM-context}
To better generalize to novel samples, a series of solutions exploit semantic context as additional cues for processing these samples. In these solutions, the semantic context enriches the model's comprehension of the novel samples by providing visual semantic information that aids in inference. 

Considering the extensive world knowledge embedded in LLMs and the convenience of querying, several methods~\cite{menon2022visual,yan2023learning,dai2023exploring,ren2024chatgpt,li2024zero,jia2023generating,yousaf2023videoprompter}  use LLMs as a source of semantic context. 
Menon and Vondrick~\cite{menon2022visual} make the first attempt to use language descriptions as external semantic contexts for visual recognition. Specifically, they query an LLM with rich world knowledge to obtain descriptive features of classes and use these class descriptors for classification via CLIP. This modification not only improves classification accuracy but also enhances the interpretability of CLIP's decisions. 

Nevertheless, descriptions generated by an LLM often contain low discriminative information and noise. Strategies for selecting or generating concise descriptions have become a hot topic of subsequent research~\cite{yan2023learning,dai2023exploring,ren2024chatgpt}. 
Yan et al.~\cite{yan2023learning} propose a novel learning-to-search method that discovers concise sets of attributes from the original generation while maintaining classification performance. Dai et al.~\cite{dai2023exploring} address the problem of unfaithful generated descriptors caused by LLM hallucinations. Their method determines when to trust the outputs of an LLM by estimating confidence scores using consistency-based uncertainty calibration. 
Meanwhile, Ren et al.~\cite{ren2024chatgpt} address the problem that descriptions of similar but different categories are indistinguishable when generating results for each category separately. They use ChatGPT to compare and group categories, and build a category hierarchy through hierarchical comparison, making the decision boundary of each category more compact.

In addition to image classification, the paradigm of deriving semantic context from an LLM has also been extended to relationship detection~\cite{li2024zero} and action recognition~\cite{jia2023generating,yousaf2023videoprompter,shi2024commonsense}. Li et al.~\cite{li2024zero} design a procedure to  decompose each predicate category into subject, object, and spatial components, and then query ChatGPT to generate descriptions for each component to construct composite descriptions as new prompts. This helps CLIP distinguish different fine-grained relation types. In the video domain, Jia et al.~\cite{jia2023generating} identify 12 pivotal attributes related to the scene, actor and body aspects. Based on these attributes, GPT-4 hierarchically generates knowledge-rich descriptions to form action-conditioned prompts. 
Shi et al.~\cite{shi2024commonsense} enhance the semantics of action categories by using BERT~\cite{devlin2018bert} to collect text proposals containing language descriptions of actions. These proposals form an action knowledge base that allows CLIP to extract action semantics by calculating similarities between text proposals and video frames. Besides enriching the semantic context for class label representations, Yousaf et al.~\cite{yousaf2023videoprompter} propose converting videos into captions and then using ChatGPT to convert these captions into language attributes and descriptions. These descriptive visual cues are integrated with visual embeddings to provide additional semantic context for various downstream tasks.

For the above methods that extract semantic context from an LLM to address the generalization challenge, we summarize their performance on the open-vocabulary image classification task and compare them with the method that solely uses CLIP for classification. Six image classification benchmarks are commonly used to evaluate the effectiveness of methods in open-vocabulary image classification: ImageNet~\cite{deng2009imagenet}, CUB~\cite{wah2011caltech}, Food101~\cite{bossard2014food}, Place365~\cite{zhou2017places}, Oxford Pets~\cite{parkhi2012cats} and Describable Textures~\cite{cimpoi2014describing}. These benchmarks differ in their class coverage. Specifically, ImageNet contains over 14 million images across approximately 220,000 daily object classes; CUB focuses on fine-grained classification with 11,788 images across 200 bird species; Food101 contains 101,000 images of 101 food classes; Place365 has more than 180,000 images spanning 365 scene classes; Oxford Pets contains 3,680 images of 37 cat and dog breeds; Describable Texture contains 5,640 images across 47 texture and pattern classes. In these datasets, the classification accuracy is used as the evaluation metric. Table~\ref{tab:class} reports the comparison results of methods~\cite{zareian2021open,menon2022visual,yan2023learning,ren2024chatgpt} using ViT-B/32~\cite{dosovitskiy2020image} as the backbone model on the aforementioned six image classification benchmarks. The VCD~\cite{menon2022visual}, LCDAtt~\cite{yan2023learning}, and CPHC~\cite{ren2024chatgpt} methods all outperform the base CLIP model, demonstrating the advantage of extracting semantic context from an LLM in enhancing the model's generalization to novel samples.

\subsubsection{Distilling teacher knowledge from a VLM}
Since the knowledge contained in a VLM is much more extensive than that contained in a close-set model trained on downstream datasets, the VLM can be regarded as the teacher model for the close-set trained student model. By distilling the useful teacher knowledge in the VLM into the student model, the student model can acquire the ability to handle novel samples. The main focus of this knowledge distillation process is  how to effectively represent and distill the teacher knowledge.

Gu et al.~\cite{gu2022open} make the first attempt to distill the teacher knowledge from an open-vocabulary image classification model~\cite{radford2021learning} to enable open-vocabulary object detection. This is achieved by aligning the region embeddings of the detected boxes with the text and image embeddings inferred by the teacher classification model. 
In addition to  distilling individual region embeddings, Wu et al.~\cite{wu2023aligning} propose  aligning the embeddings of a bag of regions to encourage the student model to understand the scene comprehensively rather than just focusing  on isolated objects. 
This method treats the bag of regions as a bag of words to obtain the bag-of-regions embeddings, which are then aligned with corresponding visual features from the CLIP image encoder. 
Similarly, Wang et al.~\cite{wang2023object} introduce a Distillation Pyramid mechanism to supplement the missing relational information in object distillation. 

Apart from distilling knowledge through aligning embeddings, Gao et al.~\cite{gao2022open} exploit the localization capability of the teacher model ALBEF~\cite{ALBEF} to generate pseudo bounding-box labels from image-caption pairs, which can be used as training data for the student detector. Kuo et al.~\cite{kuo2023open} observe that CLIP preserves the local sensitivity for detection, and thus adopt a more straightforward approach for distillation by directly incorporating the teacher model into a new student model. Specifically, they train a detector head on top of the frozen teacher CLIP, and obtain the final inference results  from a combination of detection scores and CLIP predictions. 

For open-set recognition, Liao et al.~\cite{liao2022cohoz} first build a common and scalable semantic hierarchy with all downstream datasets aligned. They prompt-tune a VLM to first recognize unknown classes and then prompt the fine-tuned VLM with handcrafted prompts for zero-shot classification. To mitigate the label bias, Liao et al.~\cite{liao2023m} propose a prompt tuning method that introduces open words from WordNet, thereby extending the prompt texts from close-set label words to more. In this way, the prompts are tuned in a simulated open-set scenario. Qu et al.~\cite{qu2024lmc} propose to simulate additional virtual open-set classes and use VLMs to generate images for both the closed-set and simulated open-set classes. These images serve as references in the inference phase of VLMs, in order to reduce models' reliance on spurious-discriminative features.

For the above methods that distill teacher knowledge from a VLM to address the generalization challenge, we summarize their performance on the open-vocabulary object detection task and compare them with two baselines: Supervised-RFS~\cite{gupta2019lvis}, which is trained using both novel and base labels, and OVR-CNN~\cite{zareian2021open}, a caption-enhanced baseline that does not use any VLM as a knowledge provider. For open vocabulary object detection methods, COCO~\cite{lin2014microsoft} and LVIS~\cite{gupta2019lvis} are widely used benchmarks. Following the common setting~\cite{bansal2018zero}, the training set of COCO is divided into a base set with 48 base classes and a target set with 17 novel classes. In total, COCO contains 107,761 training images with 665,387 bounding box annotations for base classes and 4,836 test images with 28,538 bounding box annotations for both base and novel categories. In LVIS, 866 frequent and common categories are treated as base categories, while 366 rare ones serve as novel categories. To evaluate a method’s ability to detect novel objects, COCO uses the average precision of novel categories (\textit{i.e.} Novel AP) as the main evaluation metric, while LVIS uses the average precision of rare categories (\textit{i.e.} Ap$_r$). 
Table~\ref{tab:object} shows the comparison results of methods~\cite{gu2022open,gao2022open,kuo2023open,wu2023aligning,wang2023object} using ResNet-50~\cite{he2016deep} as the backbone model on COCO and LVIS. 

As shown in Table~\ref{tab:object}, the ViLD~\cite{gu2022open}, PB-OVD~\cite{gao2022open}, FVLM~\cite{kuo2023open} BARON~\cite{wu2023aligning}, and OADP~\cite{wang2023object} methods all outperform the baselines on the novel classes while maintaining stable or even higher performance on the base classes and overall performance. These results demonstrate the advantage of distilling knowledge from a VLM in enhancing generalization to novel samples. 

For methods that extract semantic context from  LLMs, their inference processes on novel samples are highly interpretable, because the information in the extracted context is easy to understand. Therefore, if  errors occur during the inference process, the interpretability of the semantic context can effectively help researchers trace back the inference process and identify where the model's understanding differs from the expected outcome. This allows researchers to figure out the issue and optimize the method accordingly. However, due to VLMs’ confusion on compositional concepts, some fine-grained contexts that vary in detailed visual attributes may still be ambiguous to VLMs, which hinders VLMs from distinguishing visually similar samples. How to better utilize fine-grained contexts remains an open question.  
On the other hand, distilling knowledge from  VLMs allows the close-set student models to benefit from the rich visual and textual understanding that VLMs have learned, thereby establishing a more generalized connection between visual and textual modalities. This can significantly improve the  ability student models to handle open-vocabulary tasks without requiring extensive task-specific training data. Nevertheless, knowledge distillation inevitably leads to  a loss of teacher knowledge, which in turn reduces the generalization capability of the student models. This solution can  potentially be enhanced  by extracting additional semantic context from LLMs to compensate for the knowledge loss. The integration of two types of methods may bring new insights for mitigating the challenge of generalization to novel samples.

\subsection{Solutions to Task Diversity}

Due to the diversity of vision-language tasks in terms of input-output formats and reasoning processes, it is challenging to design a general model that can handle multiple tasks. 
In this context, designing specific models for different tasks and updating parameters through training brings a great burden to practical applications. 

\begin{figure*}[!t]
\centering
\includegraphics[width=0.9\linewidth]{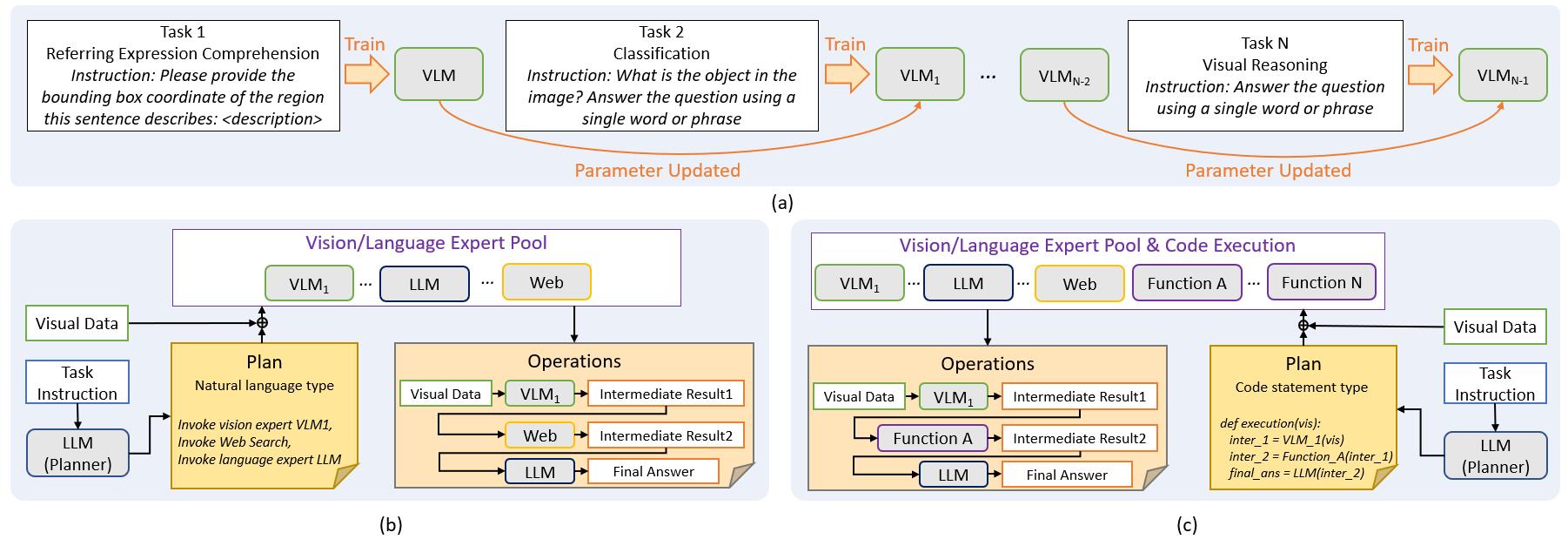}
\caption{Paradigm of using pre-trained models to conquer the task diversity challenge in vision-language tasks. (a) shows the basic idea of applying continual learning strategies on a single VLM to enable it to address different vision-language tasks. (b) and (c) show the basic idea of building a general modular system that plans with natural language and code statements to address different vision-language tasks, respectively.}
\label{fig4}
\end{figure*}

Considering the adaptability of VLMs, a new trend has emerged that focuses on continual learning of VLMs. Novel methods~\cite{qian2023decouple,chen2024coin,zheng2024beyond} of prompt learning and instruction tuning have been developed to incrementally update a VLM's knowledge. These methods allow a VLM to adapt to new vision-language tasks while retaining previously learned knowledge. Moreover, they are highly parameter-efficient, requiring only minor updates to a subset of the model's parameters or the integration of lightweight modules. This advancement marks a significant step towards enhancing VLMs' generality in addressing the growing diversity of vision-language tasks. Besides the continual learning of a single VLM, another new trend has emerged to extend traditional task-specific models into a general modular system~\cite{yang2023mm,wu2023visual,Lu2023ChameleonPC,lin2023mm,Gupta2022VisualPC,Suris2023ViperGPTVI,Choudhury2023ZeroShotVQ,Stanic2024TowardsTZ}. The core of this system is the LLM as a planner. Given a task-related instruction, the planner infers a plan consisting of a sequence of operations to be executed by various tools. 
Following the plan, the system sequentially calls different tools, including pre-trained models and pre-defined modules, to generate the final response to the input instruction. 
As shown in Fig.~\ref{fig4}, the plan can be  natural language~\cite{yang2023mm,wu2023visual,Lu2023ChameleonPC,lin2023mm} or code statements~\cite{Gupta2022VisualPC,Suris2023ViperGPTVI,Choudhury2023ZeroShotVQ,Stanic2024TowardsTZ}.
Since the pre-trained models have strong in-context learning capabilities, the general system does not require additional training process. 
With flexible planners and plug-and-play tools, such a system is not affected by task diversity and can address various visual-language tasks simultaneously, such as visual captioning, natural language image editing, factual knowledge object tagging, visual mathematical reasoning, and multi-image reasoning.

\begin{table*}[htbp]
  \centering
  \caption{A comparison of general systems proposed to address the task diversity challenge.}
   \begin{tabular}{l|cc|cc|cccc}
    \toprule
    \multicolumn{1}{c|}{\multirow{2}[4]{*}{Methods}} & \multicolumn{2}{c|}{Visual Data} & \multicolumn{2}{c|}{Planner} & \multicolumn{4}{c}{Tool} \\
\cmidrule{2-9}          & Image & Video & Planner Name & Planning Format & Experts & Web Search & Python & Tool Size \\
    \midrule
    MM-REACT\cite{yang2023mm} & $\surd$ &       & ChatGPT & natural lang & $\surd$     & $\surd$     &       & 10 \\
    Visual ChatGPT\cite{wu2023visual} & $\surd$ &       & ChatGPT & natural lang & $\surd$     &       &       & 22 \\
    Chameleon\cite{Lu2023ChameleonPC} & $\surd$ &       & GPT-4 & natural lang & $\surd$     & $\surd$     & $\surd$     & 13 \\
    VisProg\cite{Gupta2022VisualPC} & $\surd$ &       & GPT-3 & Code  & $\surd$     &       & $\surd$     & 20 \\
    ViperGPT\cite{Suris2023ViperGPTVI} & $\surd$ & $\surd$ & GPT-3 Codex & Code  & $\surd$     &       & $\surd$     & 15 \\
    MM-VID\cite{lin2023mm} &       & $\surd$ & GPT-4 & natrual lang & $\surd$     &       &       & 4 \\
    ProViQ\cite{Choudhury2023ZeroShotVQ} &       & $\surd$ & ChatGPT & Code  & $\surd$     &       & $\surd$     & 9 \\
    ZS-CVR\cite{Stanic2024TowardsTZ} & $\surd$ & $\surd$ & PaLM2 Code-bison & Code  & $\surd$     &       & $\surd$     & 14 \\
    \bottomrule
    \end{tabular}%
  \label{tab:task}%
\end{table*}%

\subsubsection{Continual Learning}
Given the extensive knowledge storage, VLMs serve as strong starting points for adapting to various vision-language tasks. Moreover, continual learning techniques can alleviate the issue of catastrophic forgetting during VLM adaptation, thereby allowing VLMs to adapt to new tasks while maintaining proficiency in previous tasks. To enable VLMs to handle various types of VQA tasks, Qian et al.~\cite{qian2023decouple} propose a multi-modal prompt learning method for VLMs, which consists of decoupled prompts and prompt interaction strategies to capture the complex interactions between visual and textual modalities. When new images or new question types that require different reasoning processes appear, the proposed method updates its decoupled prompts to learn how to perform new tasks while maintaining the capabilities of solving previous tasks. Chen et al.~\cite{chen2024coin} introduce COIN, a novel benchmark designed to evaluate VLMs in continual instruction tuning. COIN consists of 10 datasets spanning eight different tasks, including referring expression comprehension, classification, knowledge-grounded image question answering, etc. They identify that catastrophic forgetting in COIN primarily arises from VLMs' failures in intention alignment, which can be mitigated by introducing MoELoRA to use experts to acquire task-specific knowledge. Besides catastrophic forgetting, Zheng et al.~\cite{zheng2024beyond} discover a negative forward transfer in VLMs, where learning new tasks leads to performance degradation on unseen tasks. To achieve positive forward transfer, they propose to reuse pre-trained knowledge and project prompt gradients to the pre-trained space. Moreover, they allocate different subspaces for each task and project prompt gradients to the residual space to achieve anti-forgetting. These advancements highlight the effectiveness of continual learning in addressing the growing diversity of vision-language tasks, enabling a VLM to maintain adaptability and robustness across a wide range of tasks. 

\subsubsection{Planning with natural language}
Inspired by ChatGPT's ability to handle NLP problems across multiple domains, several works~\cite{yang2023mm,wu2023visual} integrate visual experts with ChatGPT to create a system capable of performing a wide range of visual-language tasks. In this system, ChatGPT acts as a planner, outputting a plan in natural language format to call other tools. 

Specifically, Yang et al.~\cite{yang2023mm} inject the usage knowledge of visual experts into ChatGPT by adding instructions about the  capabilities of each expert and providing some contextual examples for each expert within the prompts. Meanwhile, in order to facilitate ChatGPT in calling various expert models, it is also instructed to generate special watchwords when a visual expert is  needed to interpret the visual input. The output of the expert is serialized and combined with the historical message to further activate ChatGPT until no more experts are needed. The system then returns the final answer to the user. Following this paradigm, the proposed system can tackle challenging visual understanding tasks such as visual mathematical and textual reasoning, open-world concept understanding, visual planning and prediction, and so on. 
Similarly, Wu et al.~\cite{wu2023visual} introduce a new prompt manager that converts visual content, historical intermediates, and information from visual experts into understandable prompts for ChatGPT. 

In addition to incorporating off-the-shelf vision models as plug-in-play modules, Lu et al.~\cite{Lu2023ChameleonPC} also integrate web search engines and some heuristic-based modules, such as Bing Search and OpenAI tools, into their system, enhancing the reasoning capabilities of the system, including knowledge retrieval, web search, mathematical reasoning, and table understanding. As a result, the range of applications that can be achieved by this system has been expanded. 

The success of the above methods in the image domain encourages the extension of this paradigm to the video domain. Lin et al.~\cite{lin2023mm} design a system that takes a video file as input and outputs a script describing the video content. This script enables an LLM to gain a generalized understanding of the video and perform various video tasks based on the user's questions.

\subsubsection{Planning with code statements}

As a highly structured language, code is more precise than natural language. Code statements can be directly executed by machines in sequence, including conditional statements, arithmetic operations, and defined functions. Furthermore, the result returned by the previous function can be used as the input of the subsequent function to achieve information transmission. 
Given this insight, Gupta et al.~\cite{Gupta2022VisualPC} and Suris et al.~\cite{Suris2023ViperGPTVI} propose using Python programs generated from text queries to call other visual experts to process visual inputs. These methods benefit from the latest code generation LLMs (\textit{e.g.} GPT-3 Codex~\cite{chen2021evaluating}) or LLMs (\textit{e.g.} GPT-3~\cite{brown2020language}), enabling efficient program generation beyond manually created programs. 
The inclusion of built-in Python logic and mathematical operators further improves the logic of the system. 

Choudhury et al.~\cite{Choudhury2023ZeroShotVQ} also adopt the paradigm of invoking other modules through Python programs to design their system. For the input video data, the system performs reasoning at multiple semantic levels, incorporating information from individual frames, disjoint video clips, and the entire video. This capability is supported by image-based modules such as object detection and image QA and video-based modules such as video retrieval, video captioning, speech transcription, tracking and video summarization. 
Considering that video reasoning tasks require strong spatial and temporal reasoning capabilities, Stani$\acute{c}$ et al.~\cite{Stanic2024TowardsTZ} propose to divide abstract routines into spatial and temporal categories. Then, they pre-define certain operations according to prior knowledge, such as spatial routines for retrieving relations relative to a patch, and temporal routines for event localization. 

For the above methods that address the challenge of task diversity by designing general systems, we summarize them based on three aspects: the types of visual data processed by the system, the characteristics of the planner, and the tools used. 
The summarization is shown in Table~\ref{tab:task}.

A general modular system that integrates multiple pre-trained models can perform a wide range of vision-language tasks in a zero-shot or few-shot manner. This characteristic is particularly advantageous as it minimizes the need for extensive retraining or fine-tuning on new tasks. In comparison, even though continual learning avoids the massive cost of retraining VLMs, it still requires parameter optimization and storage when adapting to new tasks. Consequently, a general modular system is more efficient and scalable compared to continual learning. Moreover, by invoking corresponding APIs to execute different pre-trained models, this system further eliminates the need for locally storing model parameters and simplifies the deployment in resource-constrained environments. In addition to the computational benefits, the general modular system planning with code statements also benefited from the direct code execution of arithmetic operations and conditional statements, which are challenging for pre-trained models to perform accurately on their own. Meanwhile, the solution of designing a general system provides high interpretability through user-friendly intermediate outputs and a transparent solving process. Also, the plug-and-play configuration of both the planner and tools enhances the flexibility of the system. When a well-performing tool is newly released, it can easily replace the previous one by simply adjusting prompts and in-context examples. 
However, attempts on video data are still ongoing, where the adjustment for video characteristics requires further exploration. Video data contains much more information than a single image. For a general system, it is critical to find feasible ways to ensure  that useful information is extracted without being affected by redundant data. Moreover, for extremely long videos (\textit{e.g.} movies), an ideal system needs to represent complex video information in a more compact form to avoid exceeding the maximum acceptable length of the LLM planner.

 \section{Potential risks}
\label{sec:risk}

As discussed in the previous sections, vision-language tasks benefit greatly from pre-trained models due to their powerful advantages, such as zero-shot inference ability, knowledge implicitly stored in parameters, learned common vision-language space, and so on. Despite these well-known strengths and remarkable performance on downstream tasks, pre-trained models  also show weaknesses in certain areas, including hallucination, outdated knowledge, concept association bias, and compositional concept confusion. These issues pose potential risks when introducing pre-trained models to vision-language tasks. 
In this section, we summarize these typical issues and further discuss potential research directions where future progresses can be made.

\subsection{Hallucination}

When using pre-trained models to solve the challenges in visual-language tasks, it will introduce the hallucination problem of pre-trained models.
Specifically, the methods~\cite{yang2022empirical,tiong2022plug,wang2022language,hanu2023language,bhattacharya2023video,pan2023retrieving} that use VLMs to verbalize visual content are easily affected by the visual illusion problem of VLMs~\cite{liu2023hallusionbench}, resulting in hallucinations in visual descriptions. This hallucination arises from the architecture of generative VLMs, which typically combines a pre-trained visual encoder, an LLM, and a trainable Q-Former-like connection module. The visual component is often weaker than the language component, leading to the domination of the language module's inherent bias. As a result, VLMs tend to overly rely on language priors to answer visual questions, especially when the visual content contains out-of-domain information~\cite{dai2022plausible}. Similarly, the methods~\cite{chen2023see,zhou2023vicor,wang2023domino,you2023idealgpt,yang2023good,qi2023the,wang-etal-2023-filling,rajabzadeh2023multimodal} that use LLMs for reasoning are prone to the negative impacts of the hallucination problem in LLMs~\cite{zhang2023siren}, including input-conflicting hallucination, context-conflicting hallucinations and fact-conflicting hallucination, leading to unfaithful or nonsensical reasoning results. 
Moreover, the hallucination problem becomes more severe if the method involves chain-like interactions between pre-trained models. In this case, the hallucinations generated at any step will be accepted as  credible inputs of another model in subsequent steps, leading to a snowball-like accumulation of hallucinations~\cite{zhang2023language}.  
For example, a complex reasoning VQA sample contains an image of a girl holding cookie cutters in preparation for baking, along with a question: “What might the girl complete after a while?” Divide-and-conquer methods~\cite{chen2023see,zhou2023vicor,wang2023domino,you2023idealgpt,yang2023good,qi2023the,wang-etal-2023-filling,rajabzadeh2023multimodal} typically start by describing the visual information using a VLM . However, due to the hallucination issue, the model may incorrectly describe the scene as the girl playing with toy blocks, which diverges from the actual context. This hallucinated visual description can mislead subsequent reasoning steps and lead to cumulative errors. Specifically, these methods rely on the toy blocks as visual hints and may incorrectly infer that the girl will build a castle instead of correctly predicting that she will bake cookies. 

Despite the aforementioned negative impacts of hallucinations, current methods that introduce pre-trained models into vision-language tasks often overlook this issue and fail to implement strategies to mitigate it. As a suggestion for future work, two types of potential strategies can be incorporated to address the hallucinations arising from the use of pre-trained models in vision-language tasks.
 
The first strategy is to design hallucination mitigation techniques for pre-trained models, especially those that do not require additional training, such as general decoding strategies and prompting techniques. Most of these techniques are plug-and-play and can be easily implemented in various methods at minimal costs. For example, prompt engineering that explicitly instructs VLMs or LLMs not to spread false or unverifiable information, such as “If you don’t know the answer to a question, please don’t share false information” proposed by Touvron et al.~\cite{touvron2023llama}. In addition, simple logical consistency checks of objects in VLM responses can effectively reduce hallucinations. The second strategy involves designing appropriate verification mechanisms during the interaction between pre-trained models. Potential verification mechanisms include cross-referencing outputs from different models or integrating a feedback loop in which the outputs are assessed against the original visual inputs. Moreover, incorporating human-in-the-loop approaches, where human reviewers assess the accuracy of model predictions, can further refine the verification process. This would allow for a dynamic learning environment where the model can adjust based on real-time feedback, thereby reducing the incidence of hallucinations. These mechanisms ensure that credible results are passed to subsequent models, thus avoiding the accumulation of hallucinations.

\subsection{Outdated Knowledge}

Real-world knowledge is constantly changing and updating, but pre-trained models struggle to  keep up-to-date knowledge because they implicitly store knowledge in static parameters~\cite{wu2024continual}. Since the knowledge in the pre-trained models is limited by the timeliness of  training data, extracting knowledge from the pre-trained models to support complex reasoning in vision-language tasks may lead to  the use of outdated knowledge. 
If reasoning about complex vision-language questions requires specific knowledge and the knowledge is updated after the pre-trained model is trained, the pre-trained model can either  only provide an outdated  information or not provide relevant information at all. 
Using outdated knowledge containing incorrect or empty information as the basis for reasoning will lead to deviations in the subsequent reasoning process, ultimately resulting in wrong conclusions.  For example, a knowledge-based VQA sample contains an image of a crowd celebrating a recently awarded Nobel Prize winner, paired with a question: “What impact has the winner’s achievement had on society?” As Nobel Prize events evolve over time, pre-trained models are unable to update their internal knowledge to include newly awarded winners. Consequently, VLMs struggle to recognize these new winners, while LLMs fail to provide background information regarding the winners’ achievements. Without access to up-to-date information, methods that rely solely on static knowledge in pre-trained models struggle to associate visual content with relevant information needed to answer the question. Suffering from the outdated knowledge issue of pre-trained models, these methods fail to correctly reason about this sample. 

As solutions of the outdated knowledge issue arising from integrating LLMs into NLP tasks, knowledge editing and retrieval-augmented generation (RAG) are also practicable to the case of integrating pre-trained models into vision-language tasks. Specifically, knowledge editing updates and refines the internal knowledge of pre-trained models, ensuring that it incorporates the latest information relevant to the task. This approach either introduces an auxiliary network or modifies a subset of the model's parameters to embed new knowledge, therefore aligning the model's output with the latest information. On the other hand, RAG complements pre-trained models with up-to-date knowledge retrieved from the Internet or other real-time resources. Both visual content~\cite{yu2024visrag,sharifymoghaddam2024unirag} and text~\cite{caffagni2024wiki,qiu2024snapntell} can serve as references for retrieval, and the retrieval results are injected into the input of pre-trained models or fused with models' raw outputs. This integration of external knowledge enriches a model's context, thereby enhancing its ability to generate accurate and trustworthy responses.

\subsection{Bias of Concept Association}
\label{sec:bag}

VLMs  pre-trained using image-text contrastive objectives like CLIP are widely used to calculate  similarities between images and sentences, which serve as a common constraint to ensure the consistency between generated sentences and visual content. 
However, many studies~\cite{thrush2022winoground,yuksekgonul2022and,tang2023lemons} highlight that such VLMs tend to treat an input sentence as a bag of concepts, ignoring the syntactic structure of  sentence. As a result, relying on similarities calculated by VLMs is prone to the concept association bias, especially for the visual captioning and VQA methods that perform direct inference on test samples introduced in Sec~\ref{sec:inf-test}. 
Specifically, these methods use a VLM to fill in a masked sentence by maximizing the similarity between the filled sentence and the image, where the masked sentence is a partially generated image caption or a template derived from a visual question. 
The bag-of-words nature of VLMs makes them tend to assign higher similarity scores to sentences that encompass a wider range of elements depicted in the image. Since the masked sentences usually focus on partial concepts of the image, the filling content will ignore the sentence semantics and bias towards the missing concepts. 
For example, given an image with a purple eggplant and a yellow lemon, CLIP needs to complete the masked sentence, ``In this picture, the color of the lemon is [mask]”, by selecting a word that maximizes the visual-textual similarity score between the image and the full sentence. Since the image contains two different visual elements but the masked sentence only mentions the lemon, CLIP tends to assign a higher similarity score to filling with ``purple" than ``yellow'', as it relates to the unmentioned eggplant. Therefore, unreliable guidance of similarity scores may cause zero-shot image captioning methods to fail to describe the image. Similarly, the negative impact of concept association bias may cause zero-shot VQA methods to incorrectly answer questions like ``What color is the lemon in the image?’’ 

The negative impact of concept association bias can be mitigated from two different perspectives. From the perspective of improving VLMs, it is feasible to extend current CLIP-like VLMs by adding additional architectures that perform deeper modality interaction. This strategy goes beyond the previous shallow interaction (\textit{e.g.} a simple inner product) and enables VLMs to learn more accurate correspondences between visual and textual elements. In addition, fine-tuning on augmented data helps CLIP-like VLMs capture fine-grained differences in scenes more effectively. For example, one feasible data augmentation strategy involves constructing hard negative samples by changing the original word order of the paired image caption. 
  From the perspective of improving the integrations of VLMs, it is recommended to avoid relying solely on the visual-textual similarity calculated by a VLM as a visual constraint for sentence completion, especially when the current sentence only covers a limited number of visual elements in the image or video. Instead, as proposed by Tewel et al.~\cite{tewel2022zero}, the entire sentence should be considered as a whole and refined progressively using the similarity. Alternatively, recent advances in zero-shot image captioning~\cite{nukrai2022text,li2023decap,fei2023transferable,gu2022can,wang2023text,wang2022zero,wang2023association} learn from uni-modal data and decode sentences directly from the transformed image embeddings, thereby eliminating the reliance on similarity scores. By using the learned common vision-language space instead of similarity scores, these methods effectively mitigate the bias and outperform similarity-based methods.

\subsection{Confusion on Compositional Concepts}

Since a VLM often struggles with understanding compositional concepts~\cite{paiss2022no,paiss2023teaching,parcalabescu2022valse,ma2023crepe,ray2024cola}, 
relying on a VLM to bridge textual and visual modalities may lead to confusion in connecting  compositional concepts in texts with corresponding visual elements. 
Specifically, the differences in detailed compositional concepts among similar images or videos are easily ignored by a VLM~\cite{parcalabescu2022valse,paiss2023teaching}. 
On the other hand, given text as queries to retrieve visual content, a VLM attends to a sparse set of the input text~\cite{paiss2022no} where adjectives, numbers, and prepositions are often ignored. This makes it difficult for a VLM to accurately locate the specified visual content strictly based on the descriptive details given in the query, potentially leading to incorrect extractions of visual information. 
For example, consider a scenario where a VLM is tasked with open-vocabulary image classification on an image of a white and gray cat, likely a Ragdoll or Birman cat. To enrich VLMs’ comprehension of novel samples, many methods extract semantic context of these cat categories from an LLM. This context typically contains detailed descriptions of visual appearances, such as ``Ragdoll has large and striking blue eyes’’ and ``Birman has slightly tilted almond-shaped eyes’’, which is the key evidence to distinguish between the two categories. However, when encoding these descriptions, VLMs often overlook the differences in adjectives. This drawback hinders the VLMs' ability to classify the image correctly based on the generated context. In addition, CLIP tends to ignore fine-grained visual differences, such as color, count, and orientation, thereby encoding some different images into similar visual embeddings. For generative VLMs that take CLIP’s visual encoder as their visual components, this ambiguity in the visual encoding hinders them from correctly describing or answering questions about these visual attributes. For example, VLMs may not be able to determine whether a fruit in an image is cut or uncut, due to confusion over the subtle differences between visual embeddings of cut and uncut fruits. This drawback causes VLMs to provide inaccurate visual cues when solving complex visual questions and further mislead subsequent reasoning.

The negative impact of confusion on compositional concepts can be mitigated by improving the VLM and its integration. One feasible solution is to develop more powerful visual encoders for generative VLMs. By employing a Mixture-of-Features (MoF) strategy, we can combine CLIP's visual features with those from vision-only models trained in a self-supervised manner, thus enhancing CLIP’s ability to capture visual attributes. Moreover, some compositional concept-aware objectives~\cite{paiss2023teaching,paiss2022no} have been proposed to fine-tune VLMs, where positive and negative samples are designed based on differences of compositional concepts in images. For methods that integrate pre-trained models, it is more effective to design a unified system to comprehensively consider results from multiple models and specialized detectors rather than relying on a single VLM. This modular architecture enables different models to address specific tasks, such as object detection or optical character recognition. Furthermore, some tasks that are challenging for a VLM, such as counting objects or determining their relative positions, can be effectively represented as code statements. For example, a question like ``how many seagulls are in the image’’ can be translated into a sequence of operations: detect the seagulls and then apply a code function to sum the detected regions. By integrating code execution with the output of a VLM, such tasks can be completed more accurately and efficiently, benefiting from the strengths of both VLM's visual understanding and precise computational coding.

\section{Conclusion}
In this survey, we systematically review the  methods of integrating pre-trained models to solve challenges in vision-language tasks. 
To ensure the systematic nature of this survey, we start with an introduction to the main challenges and then categorize the existing methods according to the challenges they address, including data scarcity, escalating reasoning complexity, generalization to novel samples, and task diversity. Specifically, we summarize the paradigms into illustrations of  pipeline to provide a more intuitive understanding of  various processes involved. 
In addition, we discuss the potential risks brought by the inherent limitations of pre-trained models. We hope that this survey can provide researchers with a comprehensive overview of the current status of the field and provide insights into future research directions.

\bibliographystyle{IEEEtran}
\bibliography{IEEEabrv,reference}

\section*{Biography Section}

\vspace{-20pt}

\begin{IEEEbiographynophoto}{Yayun Qi}
received the B.S. degree from Beijing Institute of Technology (BIT), Beijing, China, in 2020. She will pursue the Ph.D. degree at the Beijing Laboratory of Intelligent Information Technology, School of Computer Science, Beijing Institute of Technology. Her research interests include computer vision, and video analysis and understanding. 
\end{IEEEbiographynophoto}

\vspace{-20pt}

\begin{IEEEbiographynophoto}{Hongxi Li}
     received the B.S. degree from Beijing Wuzi University 
 (BWU), Beijing, China, in 2022. He will pursue the M.S. degree at the Beijing Laboratory of Intelligent Information Technology, School of Computer Science, Beijing Institute of Technology. His research interests include computer vision and machine learning. 
\end{IEEEbiographynophoto}

\vspace{-20pt}

\begin{IEEEbiographynophoto}{Yiqi Song}
     received the B.S. degree from Beijing Institute of Technology (BIT), Beijing, China, in 2024. He will pursue the Ph.D. degree at the Beijing Laboratory of Intelligent Information Technology, School of Computer Science, Beijing Institute of Technology. His research interests include computer vision and video analysis and understanding. 
\end{IEEEbiographynophoto}

\vspace{-20pt}

\begin{IEEEbiographynophoto}{Xinxiao Wu (Member, IEEE)}
is a Full Professor with the School of Computer Science, Beijing Institute of Technology. She received the B.S. degree in computer science from the Nanjing University of Information Science and Technology in 2005, and the Ph.D. degree in computer science from the Beijing Institute of Technology  in 2010. From 2010 to 2011, she was a Postdoctoral Research Fellow with Nanyang Technological University, Singapore. Her research interests include machine learning, computer vision, and video analysis and understanding.
\end{IEEEbiographynophoto}

\vspace{-20pt}

\begin{IEEEbiographynophoto}{Jiebo Luo (Fellow, IEEE)}
      is the Albert Arendt Hopeman Professor of Engineering and Professor of Computer Science at the University of Rochester which he joined in 2011 after a prolific career of fifteen years at Kodak Research Laboratories. He has authored nearly 600 technical papers and holds over 90 U.S. patents. His research interests include computer vision, NLP, machine learning, data mining, computational social science, and digital health. He has been involved in numerous technical conferences, including serving as program co-chair of ACM Multimedia 2010, IEEE CVPR 2012, ACM ICMR 2016, and IEEE ICIP 2017, as well as general co-chair of ACM Multimedia 2018 and IEEE ICME 2024. He has served on the editorial boards of the IEEE Transactions on Pattern Analysis and Machine Intelligence (TPAMI), IEEE Transactions on Multimedia (TMM), IEEE Transactions on Circuits and Systems for Video Technology (TCSVT), IEEE Transactions on Big Data (TBD), ACM Transactions on Intelligent Systems and Technology (TIST), Pattern Recognition, Knowledge and Information Systems (KAIS), Machine Vision and Applications, and Intelligent Medicine. He was the Editor-in-Chief of the IEEE Transactions on Multimedia (2020-2022). Professor Luo is also a Fellow of NAI, ACM, AAAI, SPIE, and IAPR.
\end{IEEEbiographynophoto}
\vfill
\end{document}